\documentclass[letterpaper, 10pt, conference]{ieeeconf}  

\IEEEoverridecommandlockouts                              
\usepackage[utf8]{inputenc}
\usepackage[T1]{fontenc}
\usepackage{times}

\usepackage{graphicx} 
\usepackage{epsfig} 
\usepackage{amsmath} 
\usepackage{amssymb} 
\usepackage{subcaption}
\usepackage{glossaries}
\usepackage[detect-all]{siunitx}
\newacronym{fov}{FOV}{field of view}
\newacronym{dof}{DoF}{degree of freedom}
\newacronym{fps}{FPS}{frames per second}
\newacronym{imu}{IMU}{Inertial Measurement Unit}
\newacronym{sota}{SOTA}{state-of-the-art}
\newacronym{dsol}{DSOL}{Direct Sparse Odometry Lite}
\newacronym{dso}{DSO}{Direct Sparse Odometry}
\newacronym{sdso}{SDSO}{Stereo-DSO}
\newacronym{dsm}{DSM}{Direct Sparse Mapping}
\newacronym{svo}{SVO}{Semi-direct visual odometry}
\newacronym{orb}{ORB-SLAM}{ORB-SLAM}
\newacronym{pba}{PBA}{photometric bundle adjustment}
\newacronym{zncc}{ZNCC}{zero-normalized cross correlation}
\newacronym{ssd}{SSD}{sum of squared differences}
\newacronym{vo}{VO}{Visual odometry}
\newacronym{ape}{APE}{Absolute Pose Error}
\newacronym{rpe}{RPE}{Relative Pose Error}
\usepackage{xcolor}
\usepackage[breaklinks=true, colorlinks, bookmarks=true]{hyperref}
\usepackage{lipsum}

\newcommand{\dsol}{\text{DSOL}}

\newcommand\symkf{K}
\newcommand\symwin{\mathcal{K}}
\newcommand\symimg{I}
\newcommand\sympyr{\mathcal{I}}
\newcommand\symmap{\mathcal{M}}
\newcommand\symaff{\mathbf{a}}
\newcommand\symtf{\mathbf{T}}

\newcommand\symst{\mathbf{x}}
\newcommand\symuv{\mathbf{u}}

\newcommand\sympt{\mathbf{p}}

\newcommand\symid{\rho}
\newcommand\sympts{\mathcal{P}}
\newcommand\symfwd{\Pi_{\mathbf{K}}}
\newcommand\symbwd{\Pi_{\mathbf{K}}^{-1}}
\newcommand\frmworld{\mathcal{W}}
\newcommand\Hpp{\mathbf{H}_{\mathrm{pp}}}
\newcommand\Hpm{\mathbf{H}_{\mathrm{pm}}}
\newcommand\Hmp{\mathbf{H}_{\mathrm{mp}}}
\newcommand\Hmm{\mathbf{H}_{\mathrm{mm}}}

\newcommand\dxp{\delta\mathbf{x}_{\mathrm{p}}}
\newcommand\dxm{\delta\mathbf{x}_{\mathrm{m}}}
\newcommand\bpp{\mathbf{b}_{\mathrm{p}}}
\newcommand\bmm{\mathbf{b}_{\mathrm{m}}}

\title{\LARGE \bf
\dsol: A Fast Direct Sparse Odometry Scheme 
}

\author{Chao Qu, Shreyas S. Shivakumar, Ian D. Miller and Camillo J. Taylor
\thanks{We acknowledge the support of Distributed and Collaborative Intelligent Systems and Technology Collaborative Research Alliance. Ian Miller acknowledges the support of a NASA Space Technology Research Fellowship.}
\thanks{C. Qu, S. S. Shivakumar, I. Miller and C. J. Taylor are with the GRASP Laboratory, School of Engineering and Applied Sciences,
University of Pennsylvania
{\tt\small \{quchao, sshreyas, iandm, cjtaylor\} @seas.upenn.edu}}%
}
\begin{document}

\maketitle
\thispagestyle{empty}
\pagestyle{empty}

\begin{abstract}
In this paper, we describe \gls{dsol}, an improved version of \gls{dso}~\cite{Engel2018DirectSO}.
We propose several algorithmic and implementation enhancements which speed up computation by a significant factor (on average 5x) even on resource constrained platforms. 
The increase in speed allows us to process images at higher frame rates, which in turn provides better results on rapid motions.
Our open-source implementation is available at \url{https://github.com/versatran01/dsol}.
\end{abstract}


\section{Introduction}

Localization and mapping are key components of many robotic systems. 
In this work we are motivated by the requirements of micro aerial vehicles where payload and power limit the choice of sensors and the computational capacity of the system. 
For these platforms, vision is an attractive sensing modality since cameras are relatively inexpensive in terms of mass and power consumption.

We choose to build our work on \gls{dso} and \gls{sdso} as described in~\cite{Engel2018DirectSO, Wang2017StereoDL}.
The direct approach to visual odometry minimizes the \textit{photometric error} instead of a \textit{geometric} one.
In their work Engel et al.~provide extensive experiments which show that the direct and sparse combination offers unique advantages over state-of-the-art feature-based methods like ORB-SLAM~\cite{MurArtal2015ORBSLAMAV} and semi-dense direct methods such as LSD-SLAM~\cite{Engel2014LSDSLAMLD} in terms of accuracy and speed. 


The computational advantage is particularly important in the aerial context that motivates this work, where agile robots often travel up to 10m/s~\cite{Mohta2018FastAF}. 
We show that the proposed approach can run at rates of up to 500Hz on common computational hardware. 
For direct methods, the improvement in runtime also provides concomitant improvements in accuracy and robustness~\cite{Handa2012RealTimeCT}: 
faster processing times allows for higher frame rates which can then estimate moderate motions with greater fidelity and aggressive motions with higher robustness.

For many robotics application, \gls{dso} is not suitable since it is a monocular method which does not have a consistent, metric scale and has delayed initialization and re-initialization.
This poses great safety concerns, especially for flying robots. 
For this reason, we design our system to be directly initializable from stereo and/or depth images.
In this work, we focus on the stereo version and evaluate our system against \gls{sdso}, which is itself an improvement upon \gls{dso}.

The remaining sections describe the proposed system in more details, highlighting differences compared to \gls{dso} and \gls{sdso}, and explaining relevant implementation choices that lead to improvements. 
We summarize our changes as follow:
(1) we adopt the inverse compositional alignment approach in frame tracking,
(2) we track the new image w.r.t the entire window instead of the last keyframe,
(3) we propose a better stereo photometric bundle adjustment formulation compared to \gls{sdso},
(4) we greatly simplify the keyframe creation and removal criteria from \gls{dso}, and (5) we parallelize our system to effectively utilize all available computational resources.
Together, these changes lead to a simple and lightning-fast direct sparse odometry system, which we call Direct Sparse Odometry Lite (DSOL).


\begin{figure}
\vspace{6px}
\begin{subfigure}[b]{0.325\linewidth}
    \centering
    \includegraphics[width=\linewidth]{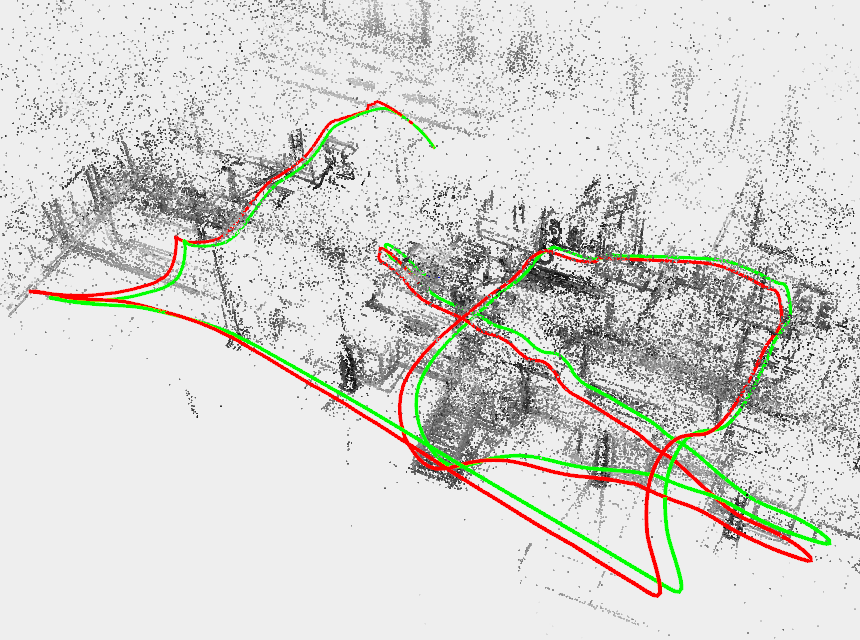}
\end{subfigure}
\begin{subfigure}[b]{0.325\linewidth}
    \centering
    \includegraphics[width=\linewidth]{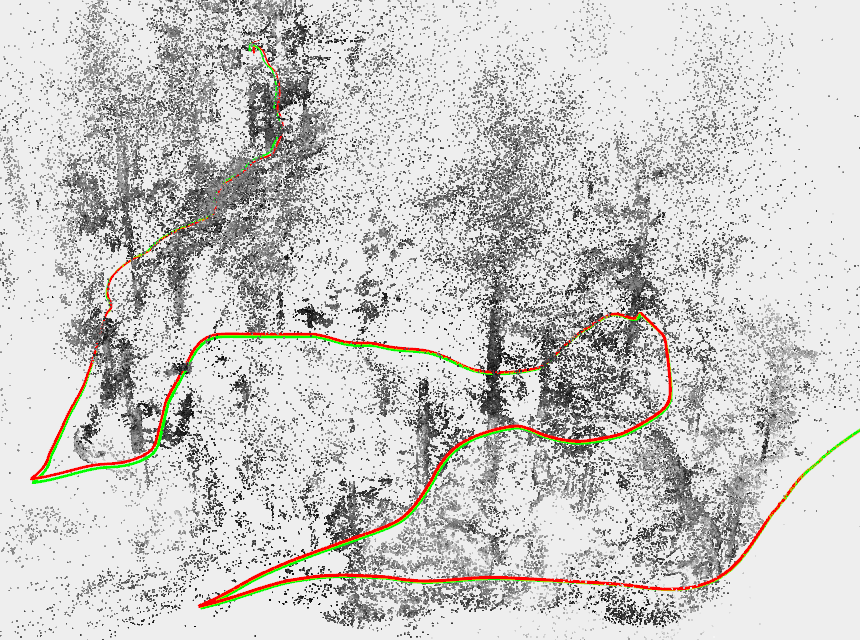}
\end{subfigure}
\begin{subfigure}[b]{0.325\linewidth}
    \centering
    \includegraphics[width=\linewidth]{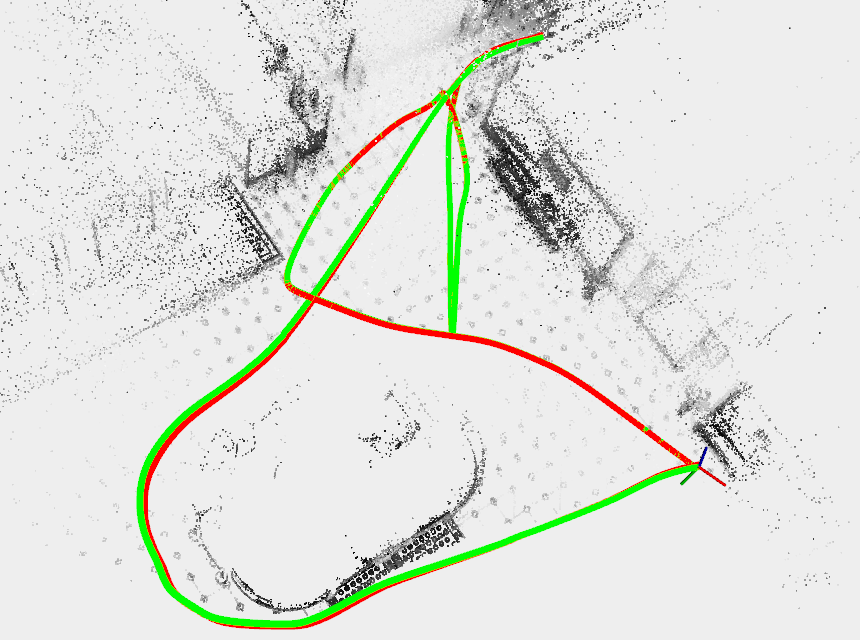}
\end{subfigure}
\begin{subfigure}[b]{0.325\linewidth}
    \centering
    \includegraphics[width=\linewidth]{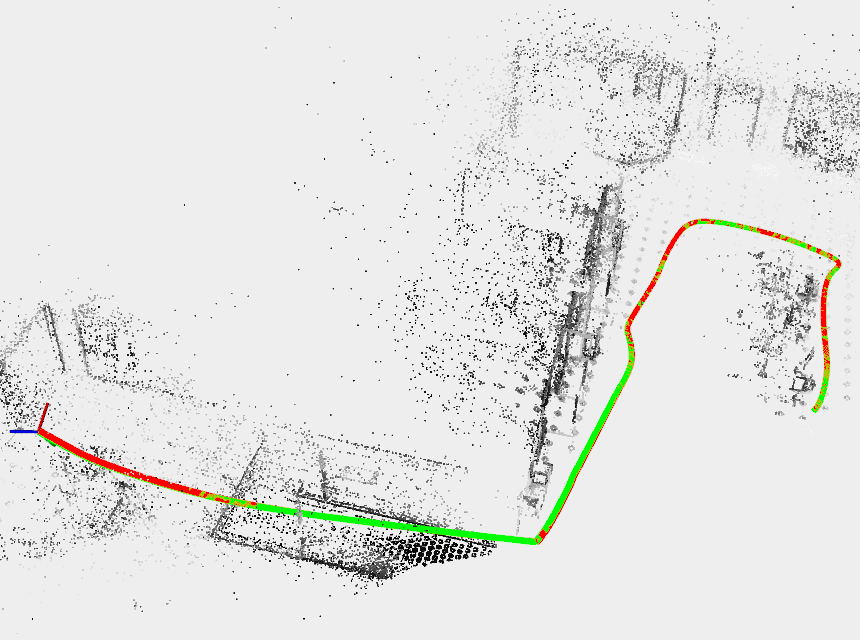}
\end{subfigure}
\begin{subfigure}[b]{0.325\linewidth}
    \centering
    \includegraphics[width=\linewidth]{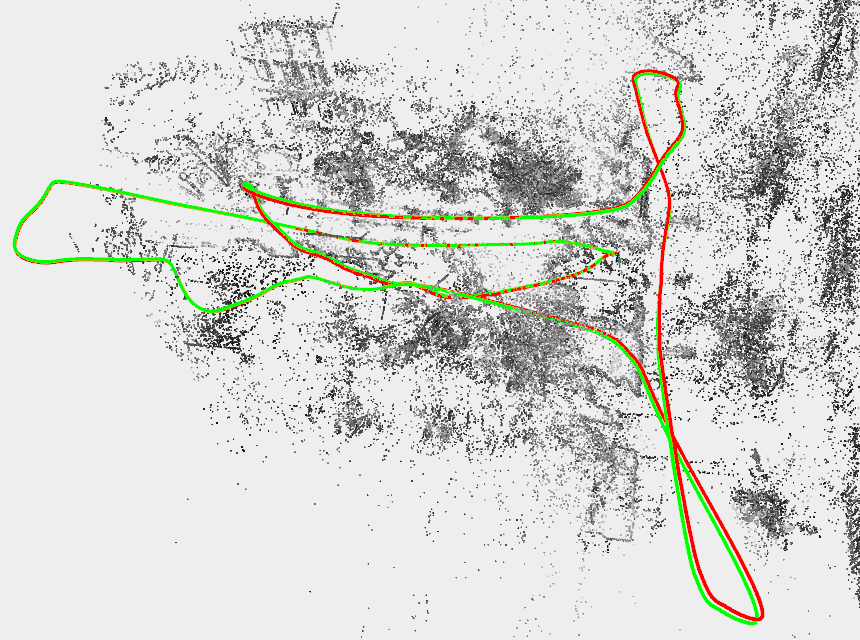}
\end{subfigure}
\begin{subfigure}[b]{0.325\linewidth}
    \centering
    \includegraphics[width=\linewidth]{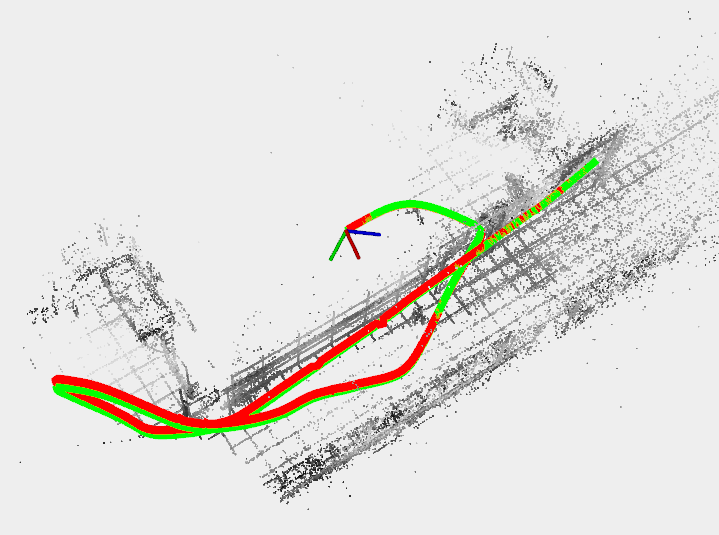}
\end{subfigure}
\caption{Results of DSOL on selected sequences from the TartanAir dataset~\cite{Wang2020TartanAirAD}.
Green lines are groundtruth trajectories, red lines are estimated trajectories.
}
\label{fig:result/tta_dsol2}
\vspace{-5mm}
\end{figure}

\section{Related Work}

\gls{vo} algorithms can be broadly categorized along the following two axes: \textbf{direct} vs. \textbf{indirect} and \textbf{dense} vs. \textbf{sparse}.
Direct methods recover model parameters directly from images by minimizing photometric error based on the brightness constancy assumption~\cite{Irani1999AllAD}.
This is in stark contrast to indirect methods, often called \textit{feature-based} methods, where correspondences are first established based on some intermediate representations, and the model parameters are optimized by minimizing reprojection errors.
Dense methods aim to use all information from the image for better accuracy and robustness at the cost of increased computation.
Sparse methods, on the other hand, recognize that image data is highly redundant and choose to only process a selected yet informative subset of the image.

Early \gls{vo}/V-SLAM systems were mostly sparse and indirect~\cite{Davison2007MonoSLAMRS, Klein2007ParallelTA, Strasdat2010ScaleDL, Strasdat2011DoubleWO}. 
This was partly due to the computation limit at the time, but also dictated by the needs of loop closure schemes in full-fledged SLAM systems, which often relied on feature descriptors~\cite{Cummins2008FABMAPPL}.
Among these systems ORB-SLAM~\cite{MurArtal2015ORBSLAMAV} stood out as a reference implementation of indirect approaches for its superior accuracy and versatility.

Dense and direct methods started gaining traction
with the advent of more powerful compute systems. 
DTAM~\cite{Newcombe2011DTAMDT} was the first system to demonstrate real-time camera tracking and dense scene reconstruction running on a GPU.
Several semi-dense systems~\cite{Engel2013SemidenseVO,Engel2014LSDSLAMLD} further reduced the computational demands by only working with high-gradient pixels.

SVO~\cite{Forster2014SVOFS} adopts a hybrid approach starting with a direct image alignment process to acquire an initial motion estimate before moving on to optimizing reprojection errors of converged map points, hence the name \textit{semi-direct}.

The direct and sparse combination that we are interested in was originally introduced in~\cite{Jin2003ASA} and later popularized by \gls{dso}~\cite{Engel2018DirectSO}.
It has then been extended to work with fish-eye lenses~\cite{Matsuki2018OmnidirectionalDD}, stereo~\cite{Wang2017StereoDL} and rolling shutter cameras~\cite{Schubert2018DirectSO}, loop closure systems~\cite{Gao2018LDSODS} and integration with deep networks~\cite{Yang2020D3VODD}.

This work does not aim to introduce an entirely new system, but rather to show that with a number of optimizations and refinements, as well as thoughtful implementation choices, we can dramatically improve the performance of a direct sparse odometry system.
\section{Direct Methods for Visual Odometry}\label{sec:method}

In this section, we outline the direct methods used in our system.
Namely, direct image alignment for frame tracking and photometric bundle adjustment for joint optimization.

\subsection{Preliminary}\label{sec:method/prelim}

Our system maintains a sliding window of $N$ keyframes $\symwin = \{\symkf_1, \cdots, \symkf_N\}$. 
Each keyframe $\symkf_i$ is associated with a Gaussian pyramid of $P$ images $\sympyr_i = \{\symimg^0_i, \cdots, \symimg^P_i\}$, a set of affine brightness parameters $\symaff_i = (a_i, b_i)^\top$, a camera pose $\symtf_i^\frmworld \in \mathbb{SE}(3)$ w.r.t. the world frame $\frmworld$, and a set of $m_k$ points parameterized by inverse depths $\sympts_i = \{\symid^1_i, \cdots, \symid^m_i\}$ hosted in this keyframe. 
The total number of points in the sliding window is $M = \sum_{\symwin} m_k$. 
Note that we need an additional image pyramid and affine brightness parameters when stereo images are available, but only keep a single set of poses and points per keyframe for the left camera.

We assume that the camera system has been calibrated before use and the images undistorted such that each camera model can be described by an intrinsic matrix $\mathbf{K}$. If a stereo rig is used we assume that the images are rectified and that the extrinsic parameters are known.
Any 3D point $\sympt=(X,Y,Z)^\top$ in the camera frame can be mapped to pixel coordinate $\symuv = (u, v)^\top$ via the projection (forward) function $\symfwd:\mathbb{R}^3 \rightarrow \mathbb{R}^2$, where
\begin{equation}
\symuv 
= \symfwd(\sympt) 
= \left(f_x \frac{X}{Z} + c_x,\ f_y \frac{Y}{Z} + c_y\right)^\top.
\end{equation}
Similarly, given pixel coordinate $\symuv$ and its inverse depth $\symid$, we can recover the 3D point $\sympt$ via the inverse projection (backward) function $\symbwd$, where
\begin{equation}
\sympt 
= \symbwd(\symuv, \symid) 
= \left(\frac{u - c_x}{f_x \symid},
\ \frac{v - c_y}{f_y \symid}, \ \frac{1}{\rho} \right)^\top.
\end{equation}
When the context is clear, we simply refer to the inverse depth and its pixel coordinate as a \textit{point} $\sympt := (\symuv, \symid)$.

For photometric camera calibration, we adopt the image formation model from~\cite{Engel2016APC}, which consists of a non-linear response function and a lens attenuation (vignetting) function.
These parameters are acquired via the software provided in ~\cite{Bergmann2018OnlinePC}.
Finally, we assume a well-calibrated camera and do not optimize any calibration parameters online.

\subsection{Frame Tracking}\label{sec:method/track}

\begin{figure}
\vspace{6px}
\begin{subfigure}[b]{\linewidth}
    \centering
    \includegraphics[width=\linewidth]{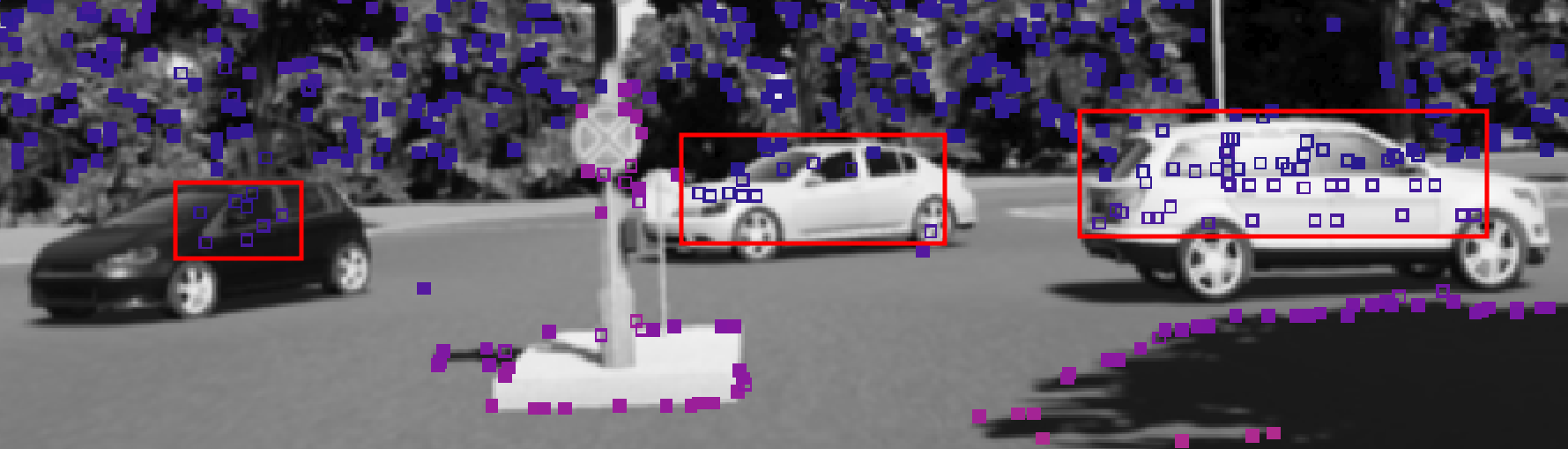}
\end{subfigure}
\begin{subfigure}[b]{\linewidth}
    \centering
    \includegraphics[width=\linewidth]{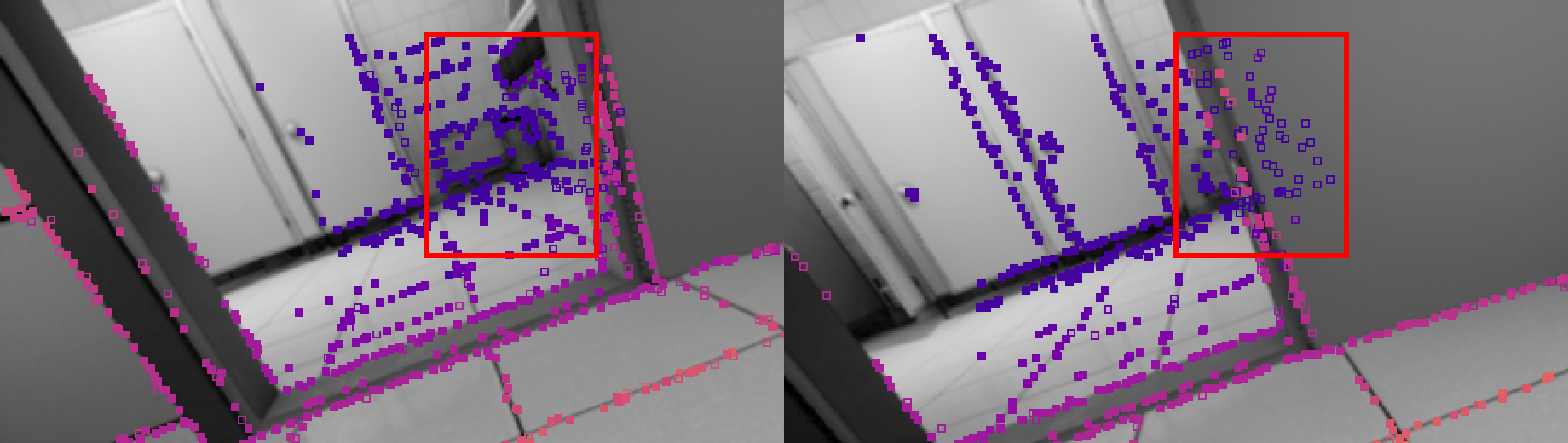}
\end{subfigure}
\caption{\textbf{Outlier rejection.}
In this visualization solid squares denote inlier feature points, while hollow ones are outliers.
The top image shows outliers on moving objects.
The bottom image shows outliers due to occlusion.
Better viewed in color and zoomed in.
}
\label{fig:method/outlier}
\vspace{-3mm}
\end{figure}

\begin{figure}
\centering
\includegraphics[width=0.5\linewidth]{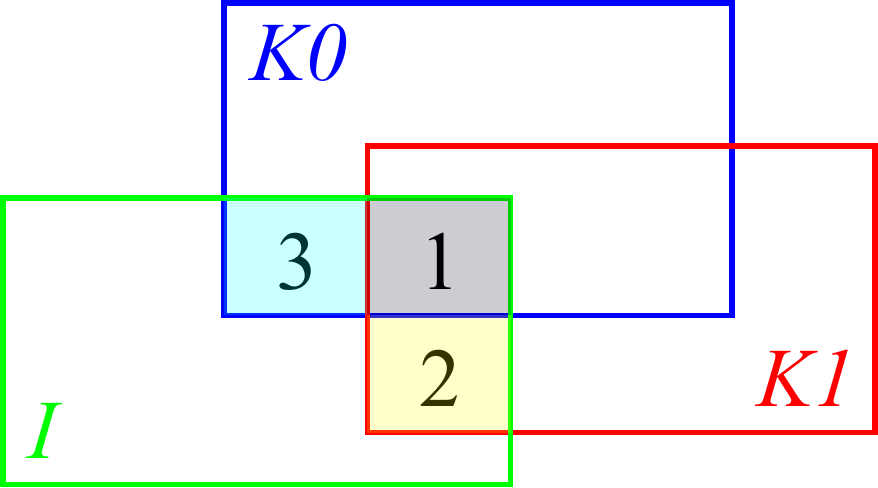}
\caption{Tracking a new image $\symimg$ w.r.t. the latest keyframe ($\symkf_1$ only) 
vs. the entire window ($\symkf_0$ and $\symkf_1$).
The former can only track points in area 1 and 2, while the later can also utilize points from area 3.}
\label{fig:method/track/fov}
\vspace{-5mm}
\end{figure}

\subsubsection{Direct Image Alignment}\label{sec:method/track/align}
Each new image $I_j$ is tracked w.r.t. a local map $\symmap = \{ \sympts_1, \cdots, \sympts_N \}$ via direct image alignment~\cite{Baker2004LucasKanade2Y}.
The parameters we wish to optimize are $\symst_j = (\symtf^\frmworld_j, \symaff_j)$ while all keyframe poses stay fixed.
A simple constant velocity model provides an initial guess to $\symtf^\frmworld_j$ in the absence of an \gls{imu}.

Following standard practice~\cite{Engel2018DirectSO}, 
we define the photometric error $E_\sympt$ of a point $\sympt =(\symuv, \symid) \in \sympts_i$ in a reference frame $I_i$ (a keyframe), observed in the target frame $I_j$ (the new image), as the weighted \gls{ssd} over a small patch of pixels $\mathcal{N}_\symuv$ .
Let
\begin{align}\label{eqn:method/track/cost}
E_\sympt 
&= 
\sum_{\symuv_k \in \mathcal{N}_\symuv}
w_k r_k^2(\symst_j) \\
&=
\sum_{\symuv_k \in \mathcal{N}_\symuv}
w_k 
\left( 
(\symimg_i[\symuv_k] - b_i ) 
- \frac{e^{a_i}}{e^{a_j}} (\symimg_j[\symuv'_k] - b_j )\right) ^2
\end{align}
where $r_k$ denotes the photometric residual associated with pixel $u_k$, $I[\cdot]$ is the pixel index operator and $\symuv'_k$ the warped pixel of $\symuv_k$ from $I_i$ to $I_j$, given by the warping function $W(\cdot, \cdot)$
\begin{equation}
\symuv'_k 
= W(\sympt_k, \symtf_i^j)
= \symfwd(\sympt'_k)
= \symfwd(\symtf_i^j \symbwd(\symuv_k, \symid_k)).
\end{equation}
$w_k = w_k^g \cdot w_k^r$ is a combined weighting factor where 
\begin{align}
w_k^g = \frac{c^2}{c^2 + \| \nabla I_i[\symuv_k] \|^2_2},
\quad
w_k^r = \frac{\nu + 1}{ \nu + (\frac{r_k}{\sigma_k})^2}.
\end{align}
$w_k^g$ penalizes pixels with high gradients~\cite{Engel2013SemidenseVO} and $w_k^r$
is a robust weight assuming residuals of the sparse model follow a t-distribution with \gls{dof} $\nu$~\cite{Kerl2013RobustOE, Zubizarreta2020DirectSM}.
In addition, we employ a simple gradient-based outlier rejection scheme to handle large errors due to occlusions or moving objects, as shown in Fig.~\ref{fig:method/outlier}.
An observation is considered bad if $r_k^2 > \| \nabla I_i[\symuv_k] \|^2_2$.
If more than one pixel within a patch is bad, the entire patch is discarded from the optimization.

The total error for direct image alignment is
\begin{equation}
E_{\mathrm{align}} 
= \sum_{I_i \in \symwin} \sum_{\sympt \in \sympts_i^j} E_{\sympt},
\end{equation}
where $\sympts_i^j$ is the set of points in $\symkf_i$ that are visible in $I_j$.

In contrast to DSO we adopt the \textbf{inverse compositional alignment}~\cite{Baker2004LucasKanade2Y} method, which swaps the roles of $I_i$ and $I_j$, allowing computationally demanding steps to be performed ahead of time~\cite{Klose2013EfficientCA}.
Our system also supports tracking using both stereo images.
This provides improved accuracy at the cost of extra computation.
It also improves robustness in situations where one of the cameras is temporarily blocked: map points that are still visible in the other image can create enough constraints to estimate model parameters.

\subsubsection{Frame-to-Window Alignment}\label{sec:method/track/window}
In \gls{dso}, the new image is tracked w.r.t. the last keyframe. All active points are projected into this keyframe. 
Frame parameters are then optimized using conventional two-frame direct image alignment~\cite{Lucas1981AnII}.
Although this worked well in practice, we found it to be sub-optimal for the following reasons:
(1) Tracking w.r.t. a single image limits the number of points that can be used, since
points that are visible in the new image but not in the last keyframe are discarded.
(2) Every point will be projected twice: first into the last keyframe, then into the new image.
This incurs computational overhead and prevents the reuse of previously computed gradients and Jacobians.

Instead, we propose tracking the new image w.r.t. the entire window.
By using all available keyframes, we essentially create a larger virtual \gls{fov}, 
allowing more points to be used during alignment. 
This is illustrated in Fig.~\ref{fig:method/track/fov}, 
where tracking image $\symimg$ w.r.t. both $\symkf_0$ and $\symkf_1$ incorporates more points into the optimization compared to using only $\symkf_1$.
This improves robustness and reduces the number of keyframes created in cases where the camera undergoes repetitive motion.
Note that since all keyframe poses are in the same reference frame, 
the inverse compositional formulation still applies.

\subsection{Bundle Adjustment}\label{sec:method/adjust}

\subsubsection{Photometric Bundle Adjustment}\label{sec:method/adjust/pba}

Upon adding a new keyframe, we perform windowed \textbf{\gls{pba}} to update all model parameters.
\begin{align}
\symst
&= (\symst_1, \cdots, \symst_N)^\top \\
&= (\symtf_1, \symaff_1, \sympts_1, \cdots, \symtf_N, \symaff_N, \sympts_N)^\top.
\end{align}
We use the same cost function in Eqn.~\ref{eqn:method/track/cost} with total error
\begin{equation}
E_{\mathrm{pba}} 
= \sum_{I_i \in \symwin} \sum_{I_j \in \overline{\symwin}_i} \sum_{\sympt \in \sympts_i^j} E_{\sympt},
\end{equation}
where $\overline{\symwin}_i$ is the set of keyframes excluding $K_i$.
By forgoing explicit data association between images, \gls{pba} tries to project all points from each keyframe to the others, forming a large number of implicit photometric correspondences.

\begin{figure}
\vspace{6px}
\begin{subfigure}[b]{0.32\linewidth}
    \centering
    \includegraphics[width=\linewidth]{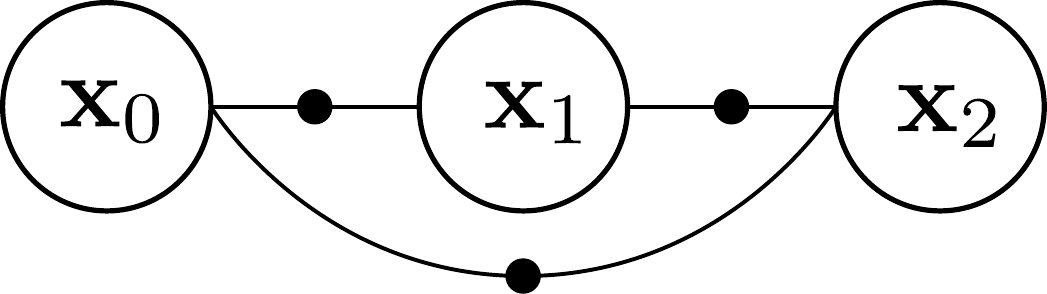}
    \caption{\gls{dso}~\cite{Engel2018DirectSO}}
    \label{fig:method/adjust/factor/dso}
\end{subfigure}
\hfill
\begin{subfigure}[b]{0.32\linewidth}
    \centering
    \includegraphics[width=\linewidth]{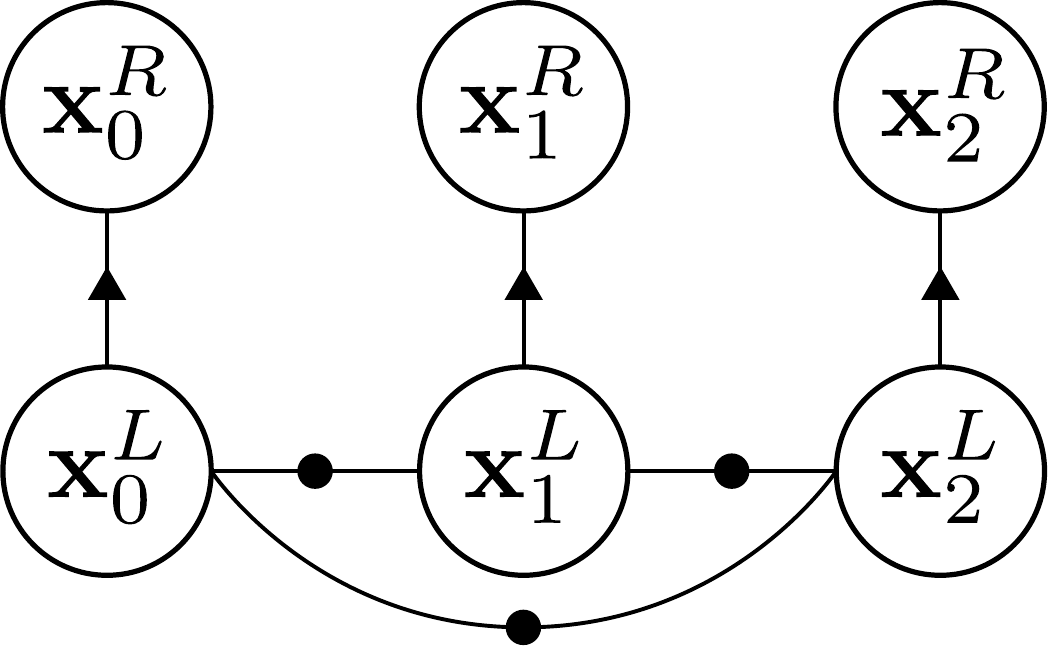}
    \caption{SDSO~\cite{Wang2017StereoDL}}
    \label{fig:method/adjust/factor/sdso}
\end{subfigure}
\hfill
\begin{subfigure}[b]{0.32\linewidth}
    \centering
    \includegraphics[width=\linewidth]{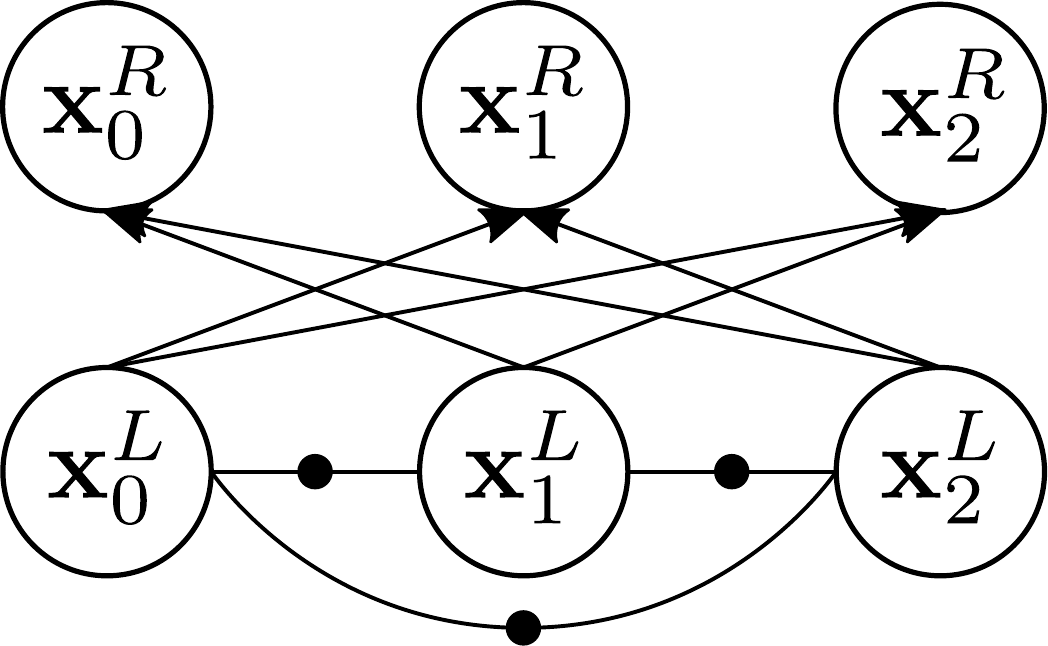}
    \caption{DSOL (\textbf{ours})}
    \label{fig:method/adjust/factor/dsol}
\end{subfigure}
\caption{
\textbf{Factor graphs} of \gls{pba} with 3 keyframes.
A directed edge in the graph highlights the source of projection.
For example, the edge $\symst_0^L\rightarrow \symst_0^R$ represents the set of factors formed by projecting $\sympts_0$ into $I_0^R$.
An undirected edge combines the pairwise projections between two keyframes.
}
\label{fig:method/adjust/factor}
\vspace{-5mm}
\end{figure}

The factor graph representation of \gls{pba} is shown in Fig.~\ref{fig:method/adjust/factor/dso}.
A directed edge indicates the source of projection and an undirected one combines projections from both keyframes.
A window of size $N$ requires $N (N -1)$ pairwise projections between keyframes.
For example, if we have 5 keyframes, each hosting 500 points with 5 pixels per patch, optimizing over 4 pyramid levels with 4 iterations per level, there will be $5\times 4 \times 500 \times 5 \times 4 \times 4 = 8\times 10^5$ total residuals.
This is several orders of magnitude more compared to indirect methods~\cite{MurArtal2017ORBSLAM2AO}, and poses significant challenges to real-time operation.
For this reason, \gls{dso} limits the total number of points to 800 in its fast setting.
However, we show that it is possible to achieve even a 100fps keyframe rate without limiting the number of points by leveraging cache-friendly data structures and a parallel implementation, which we describe in Sec.~\ref{sec:detail}.

\subsubsection{Stereo PBA}

Using stereo images with a known baseline in \gls{pba} renders the scale of the system directly observable and significantly reduces scale-drift compared to its monocular counterpart.
In \gls{sdso}~\cite{Wang2017StereoDL}, static stereo costs were added alongside temporal ones from \gls{dso}, which constrain a subset of the total parameters.
The corresponding factor graph is shown in Fig.~\ref{fig:method/adjust/factor/sdso}, where static stereo factors are depicted by direct edges and temporal ones by undirected ones.
A carefully tuned stereo coupling factor was introduced to balance the weight between these two types of cost.

In contrast, we choose to project every point into every other image (both left and right) and from an even denser network of constraints, as shown in Fig.~\ref{fig:method/adjust/factor/dsol}.
This eliminates the need for a coupling factor as all costs are now temporal, with half of them depending on the fixed baseline.
Note that this also doubles the runtime of the algorithm.
However, we believe that the time penalty is more than justified by the improved robustness, direct initialization, and scale recovery, which are indispensable to real-world applications.

\subsection{Optimization and Marginalization}\label{sec:method/optim}

We optimize both $E_{\mathrm{align}}$ and $E_{\mathrm{pba}}$ using a few iterations of the Gauss-Newton algorithm at each pyramid level. We stop early if the optimization plateaus.
The linearized system (at linearization point $\symst_0$) is given by
\begin{align}
\mathbf{H} \delta\symst = \mathbf{b} 
\quad \mathrm{or} \quad 
\mathbf{J}^\top \mathbf{W} \mathbf{J} \delta\symst
= -\mathbf{J}^\top \mathbf{W} \mathbf{r},
\end{align}
where $\mathbf{H}$ is the Hessian, $\mathbf{J}$ the Jacobian, $\mathbf{W}$ the weight matrix, and $\mathbf{r}$ the residual vector.
For direct image alignment, the linear system is typically small because the variables are parameters of the current frame $\symst_j$.
For \gls{pba} the dimension of $\mathbf{H}$ scales linearly with the size of the sliding window and the number of pixel features and is typically on the order of $10^3$.
Although large, this Hessian exhibits certain exploitable sparsity structures.
Specifically, we can divide the linear system into the following block form
\begin{align}
\begin{bmatrix}
\Hpp & \Hpm \\
\Hmp & \Hmm
\end{bmatrix}
\begin{bmatrix} \dxp \\ \dxm \end{bmatrix}
= \begin{bmatrix} \bpp \\ \bmm \end{bmatrix},
\end{align}
where subscript $\mathrm{p}$ denotes the set of pose and affine variables, and $\mathrm{m}$ denotes the set of inverse depths. 
The bottom right block $\Hmm$ is a diagonal matrix because points are parameterized by inverse depths. 
To solve the full system, we first marginalize out all depth variables by applying the Schur complement and arrive at the reduced system $\Hpp^* \dxp = \bpp^*$,
\begin{align}\label{eqn:method/optim/linear}
\Hpp^* &= \Hpp - \Hpm \Hmm^{-1} \Hmp \\
\bpp^* &= \bpp - \Hpm \Hmm^{-1} \bmm
\end{align}
$\dxp$ can be solved for efficiently and back-substituted to yield
\begin{align}
\dxm = \Hmm^{-1} (\bmm - \Hmp \dxp)
\end{align}
The new state is then updated via $\symst \leftarrow \symst \boxplus \delta \symst$, where $\boxplus$ is the box-plus operator defined in~\cite{Hertzberg2008AFF}.

When we need to remove variables from the active set (removing a keyframe $K_i$ from the window), we compute a linear system $\mathbf{H}_i$, similar to Eqn.~\ref{eqn:method/optim/linear}, containing residuals that depend on $\symst_i$ (variables to be removed).
We first marginalize all points from $\mathbf{H}_i$ and then the frame variables.
The result is a quadratic form on the rest of the keyframe parameters $\hat\symst_i$, which can be converted to a set of linear constraints $\hat{\mathbf{H}}_{i} \delta \hat{\symst}_{i} = \hat{\mathbf{b}}_{i}$.
This can then be directly added to the normal equation of subsequent optimization as a linear prior.
Note that for all keyframes that have a prior term, we need to fix their first estimates at $\overline{\symst}$ and only update the quadratic marginalization prior $\Delta \symst$ to avoid inconsistencies introduced by repeated linearizations~\cite{Li2012ConsistencyOE}. The evaluation point can then be recovered by $\symst = \overline{\symst} \boxplus (\Delta \symst + \delta \symst)$.
We refer the reader to~\cite{Engel2018DirectSO, Leutenegger2015KeyframebasedVO} for more details on marginalization.

\section{Visual Odometry Front-end}

\begin{figure}
\vspace{6px}
\centering
\begin{subfigure}[b]{0.49\linewidth}
    \centering
    \includegraphics[width=\linewidth]{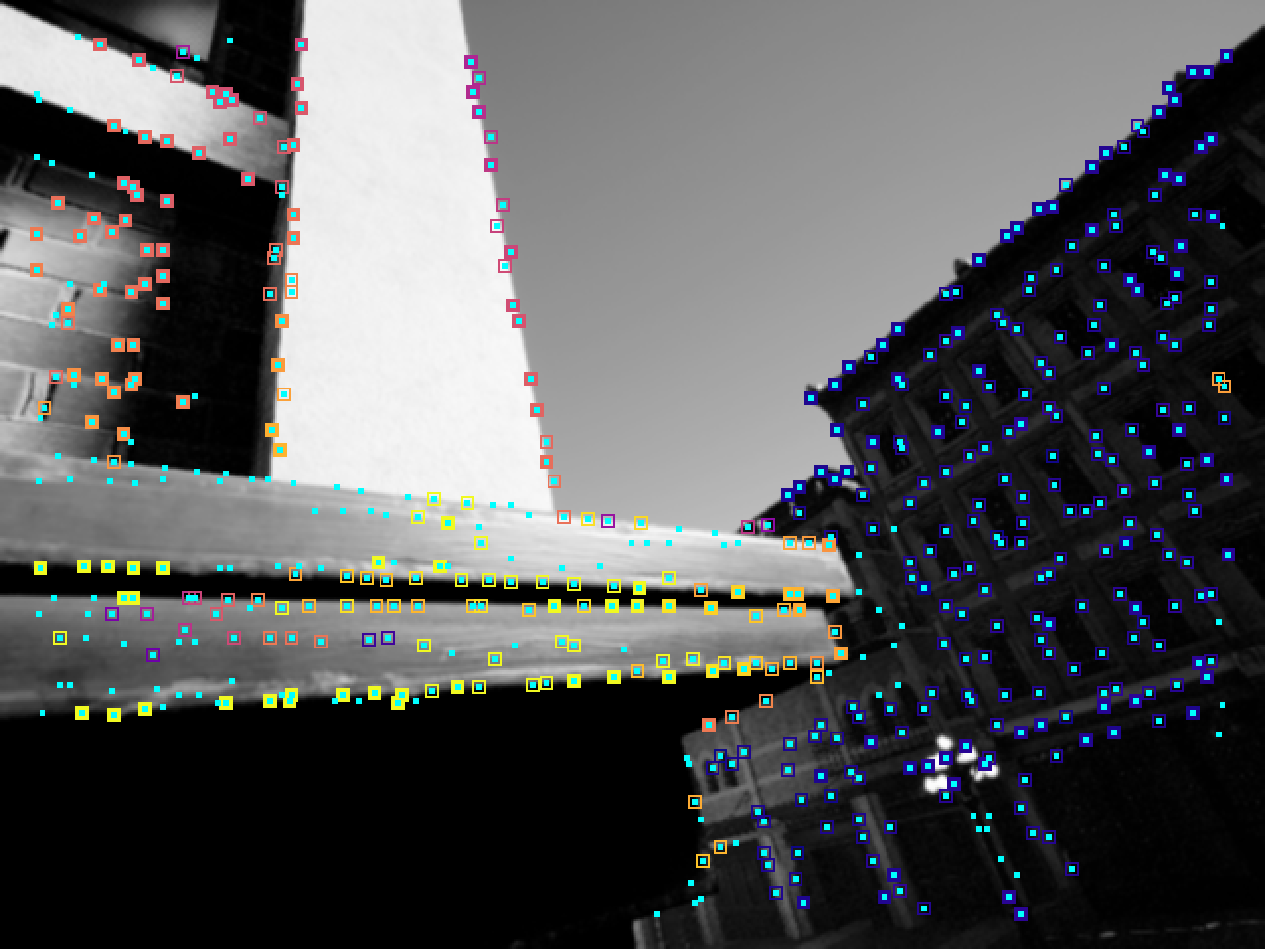}
\end{subfigure}
\hfill
\begin{subfigure}[b]{0.49\linewidth}
    \centering
    \includegraphics[width=\linewidth]{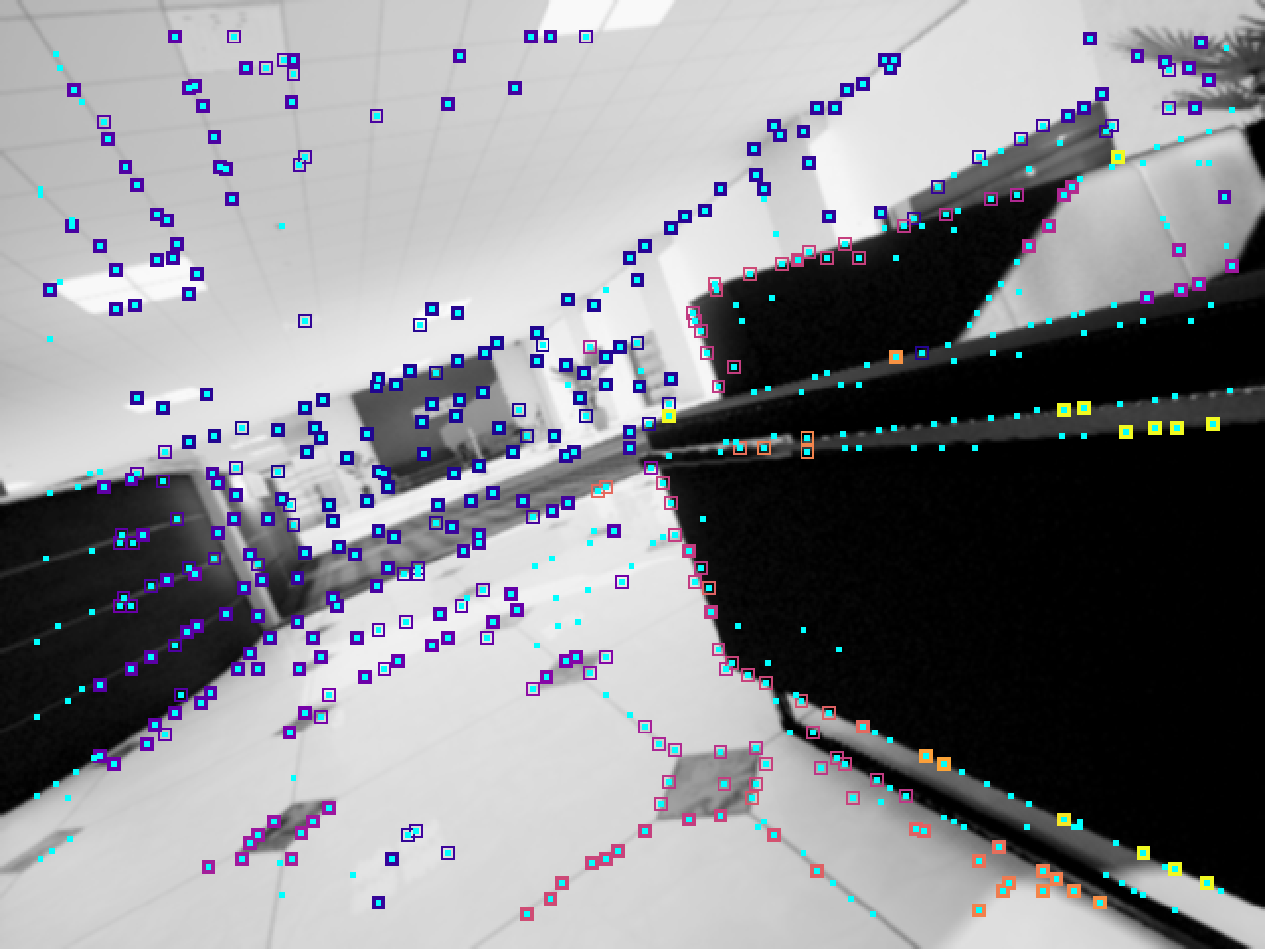}
\end{subfigure}
\caption{
\textbf{Keyframe initialization.} 
Selected points with large local gradients are in cyan.
Colored square around each pixel reflects its initial inverse depth. 
Depths are initialized from stereo matching (\textbf{thick} border) and local map (\textit{thin} border). 
Better viewed in color and zoomed.
}
\label{fig:frontend/kfinit}
\vspace{-5mm}
\end{figure}

In this section, we describe our \gls{vo} front-end, 
which manages keyframes and points within the window, including their \textbf{creation}, \textbf{initialization} and \textbf{removal}.

\subsection{Keyframe Creation}

\gls{dso} employs a complicated set of rules for keyframe creation, which involves computing the mean squared optical flow, the mean flow without rotation and the change in affine brightness factor, all from the current image to the last keyframe~\cite{Engel2018DirectSO}. 
The weighted sum of these quantities is then compared to another threshold to determine whether a new keyframe is needed, resulting in 4 extra parameters to tune.
In addition, \gls{dso} follows the philosophy as ORB-SLAM~\cite{MurArtal2015ORBSLAMAV, MurArtal2017ORBSLAM2AO}, which is to take many keyframes upfront and sparsify them with early marginalization.
However, this strategy is ineffective when computational resources are limited. 

Keyframe creation is arguably the most time-consuming part of the entire pipeline (often 5-10x slower than tracking, see Table~\ref{tab:result/runtime}).
Upon adding a new keyframe, one needs to select candidate pixels, precompute patch gradients, initialize point depths, and perform \gls{pba} and other necessary bookkeeping.
Creating too many keyframes (even in a separate thread) will cause the system to eventually lag behind the image frame rate.

We argue that a keyframe is only needed when the current image cannot be reliably tracked w.r.t. the sliding window. 
If enough points from the local map can be successfully projected into the image, then we can simply keep using these keyframes.
This prevents the addition of new ones that do not contribute very much to frame tracking.
Quantitatively, we define the tracking ratio $Q$ as the quotient between the number of tracked points and the number of selected points from all keyframes in the window.
We create a new keyframe if $Q$ falls below $Q_{\min}$.  
A higher $Q_{\min}$ therefore results in more keyframes taken.

\subsection{Keyframe Initialization}\label{sec:frontend/kfinit}

When a new keyframe is created, we select points with high local-gradients, extract a small patch around it and initialize its inverse depth if possible, followed a final \gls{pba} (Sec.~\ref{sec:method/optim}).
Once a keyframe is added it is immediately used for frame tracking (Sec.~\ref{sec:method/track}).

\subsubsection{Point Selection}\label{sec:frontend/kfinit/select}
We divide the image into a grid of cells with size 16$\times$16 pixels and select at most one point from each cell.
We start by finding one candidate pixel per cell with a large image gradient.
We then choose candidates with gradient greater than $g_{\min}$.
If this first round of selection results in too few points, we do a second round with $g_{\min} / 2$.  However, $g_{\min}$ is a running quantity, so we also decrease it by $\delta_g$ for the next image.
On the other hand, if too many are chosen, we simply increase $g_{\min}$ by $\delta_g$.
In this way $g_{\min}$ is constantly adjusted to provide the desired number of points.
Hence $g_{\min}$ is continuously adapted to select a sufficient number of points with a good spatial distribution and is upper-bounded by the total cells in the grid.

Unlike \gls{dso}, we avoid selecting points from image areas where previous map points were found. 
This is to prevent creating duplicate points that correspond to the same scene surface area, which negatively impact the linearized system by introducing unwanted dependencies~\cite{Alismail2016PhotometricBA}.
To this end, we project all map points into the current image and slightly dilate them to create a binary mask.
Only pixels outside of this mask will participate in the selection process.
Fig.~\ref{fig:frontend/kfinit} shows selected pixels for sample images.

\subsubsection{Patch Extraction}

We extract a small patch around each selected point.
We adopt a simple cross-shaped patch with a total of 5 pixels (see pattern 1 in~\cite{Engel2018DirectSO}).
This is done at every pyramid level by down-scaling the pixel coordinate accordingly. 
Intensity values and gradients are computed for each pixel within the patch using bilinear interpolation.
These quantities are re-used during tracking to reduce computation.

\subsection{Point Depth Initialization}

Points in a new keyframe have their inverse depths initialized in the following order if available: \textbf{depth image}, \textbf{stereo images}, \textbf{map points}.
Any point depth set by a previous method will not be considered by latter ones. 
Note that we do not perform small baseline stereo matching to initialize depth, as was done in \gls{dso}~\cite{Engel2018DirectSO}.
We also do not distinguish active points from candidate ones: all points with valid depths are active and can be used in tracking and \gls{pba}.

\subsubsection{From depth image}
If an aligned depth image is available (either from a depth camera, a lidar projection, or even a depth prediction network), we directly set the inverse depth of any pixel with a valid measurement.

\subsubsection{From stereo images}
If a pair of rectified stereo images is available, a simple sparse stereo matching scheme is employed on pixels that are yet to be initialized.
Since the scale pyramid of both images are already constructed, we carry out a correspondence search in a coarse-to-fine manner to reduce the number of comparisons.
In particular, we only search the max disparity range at the lowest pyramid level, and gradually refine the disparity estimate while going up.
Similar to \gls{sdso}~\cite{Wang2017StereoDL}, we use \gls{zncc} as the similarity metric on a small patch of pixels ($5\times 7$).
Note that our goal here is to initialize as many points as possible, therefore we do not enforce any strict outlier rejection scheme, but rely on \gls{pba} to remove points with large errors.

\subsubsection{From map points}
Finally, if there are points remain uninitialized, we attempt to estimate their depths from the local map.
The direct image alignment step generates a set of valid projections from all map points onto the current image, which can be used for this purpose.
We average the inverse depths of all points (observed from the new image) that fall into a single cell and use it to initialize the new point selected from this cell. 
Again, accuracy is not important here, since they will be refined or removed by \gls{pba}.
Fig.~\ref{fig:frontend/kfinit} also shows the depth initialization process for some sample images.

\subsection{Keyframe Removal}
To bound computational complexity, the window size is fixed at $N$. 
\gls{dso} proposed several heuristics for keyframe removal, which could remove multiple keyframes at the same time.
Specifically, they remove keyframes where less than $5\%$ of their points are visible in the most recent keyframe.
This was partly due to their tracking strategy, which requires projection of active points into the latest keyframe (Sec.~\ref{sec:method/track}).
Thus it is natural to remove keyframes that do not have enough overlap with the last keyframe.

We argue that this is sub-optimal, despite having a slight improvement in speed (due to fewer keyframes in the window).
Since points anchored in a particular keyframe are projected into every other keyframe in \gls{pba}, removing a keyframe weakens the links among the network of constraints formed by all pairwise projections.
Although proper marginalization does retain a linear prior connecting the remaining keyframes, it is weaker compared to the original nonlinear constraints.
Therefore we choose to remove only one keyframe at a time with the fewest points visible in the current image.
By always having a full window, we ensure strong connections among keyframes.
Finally, before actually discarding the keyframe, we perform keyframe marginalization according to Sec.~\ref{sec:method/optim}.


\section{Implementation Details}
\label{sec:detail}
\begin{figure}
\vspace{6px}
\centering
\begin{subfigure}[b]{0.32\linewidth}
\centering
\includegraphics[width=\linewidth]{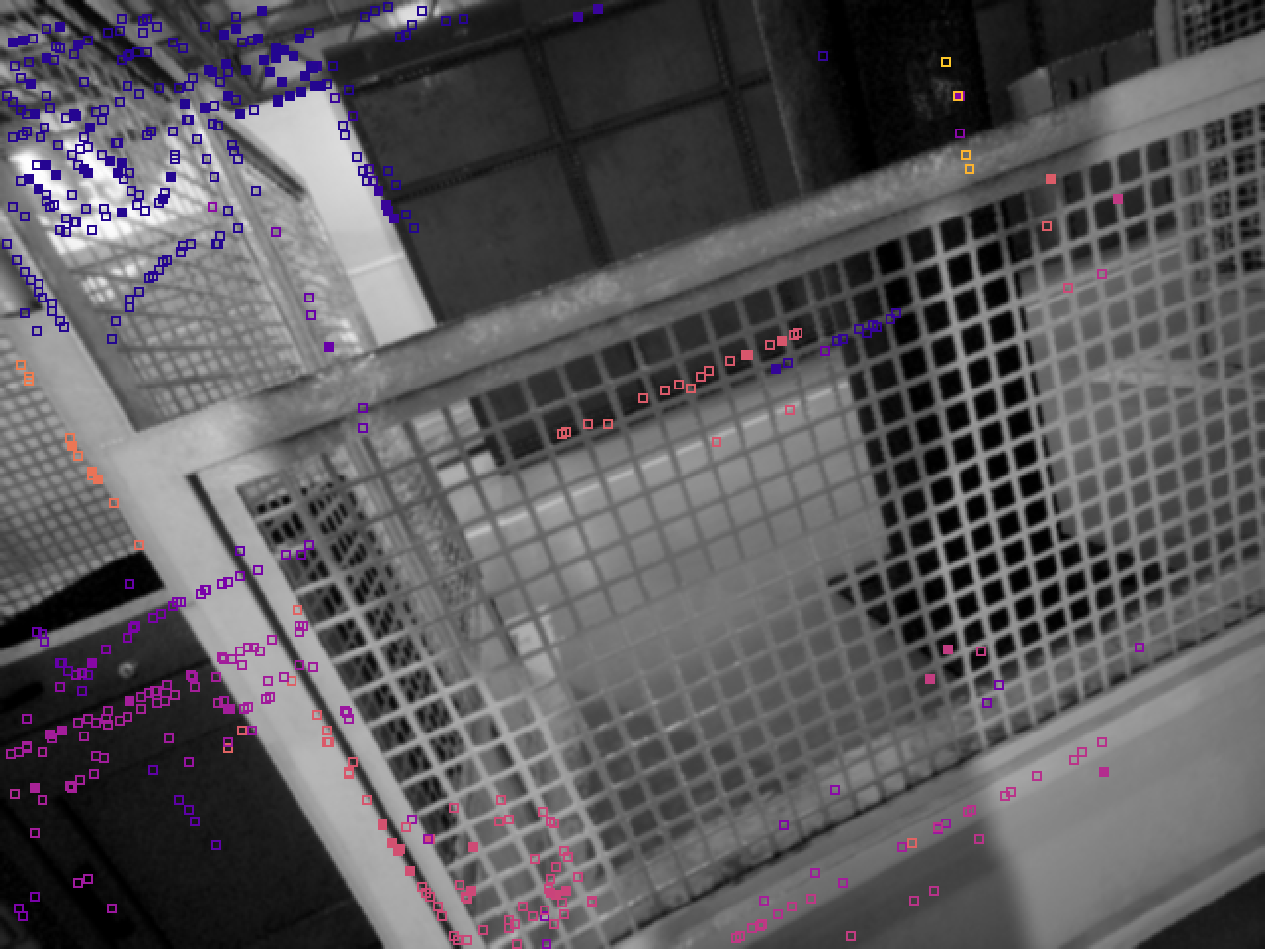}
\end{subfigure}
\hfill
\begin{subfigure}[b]{0.32\linewidth}
\centering
\includegraphics[width=\linewidth]{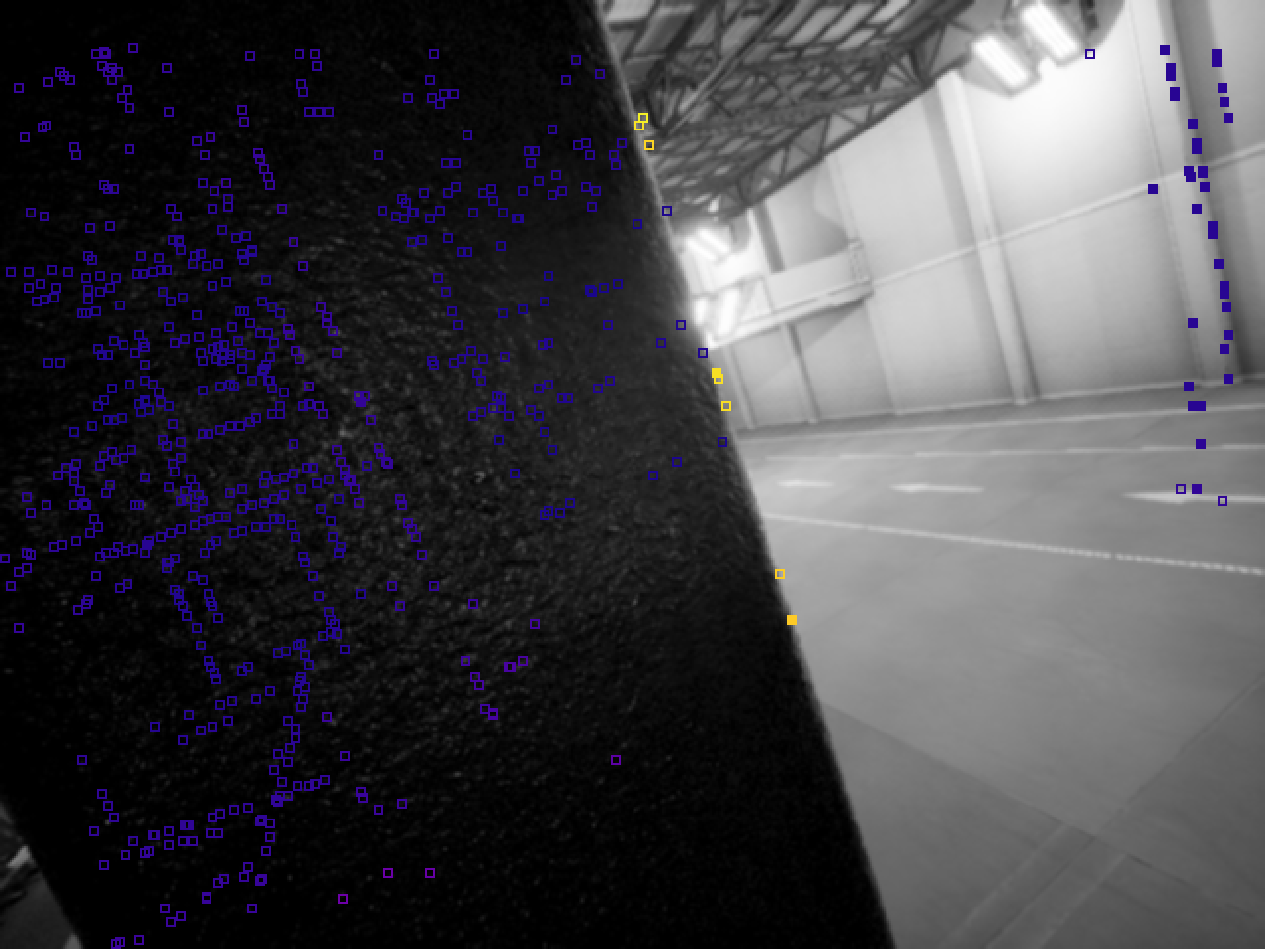}
\end{subfigure}
\hfill
\begin{subfigure}[b]{0.32\linewidth}
    \centering
    \includegraphics[width=\linewidth]{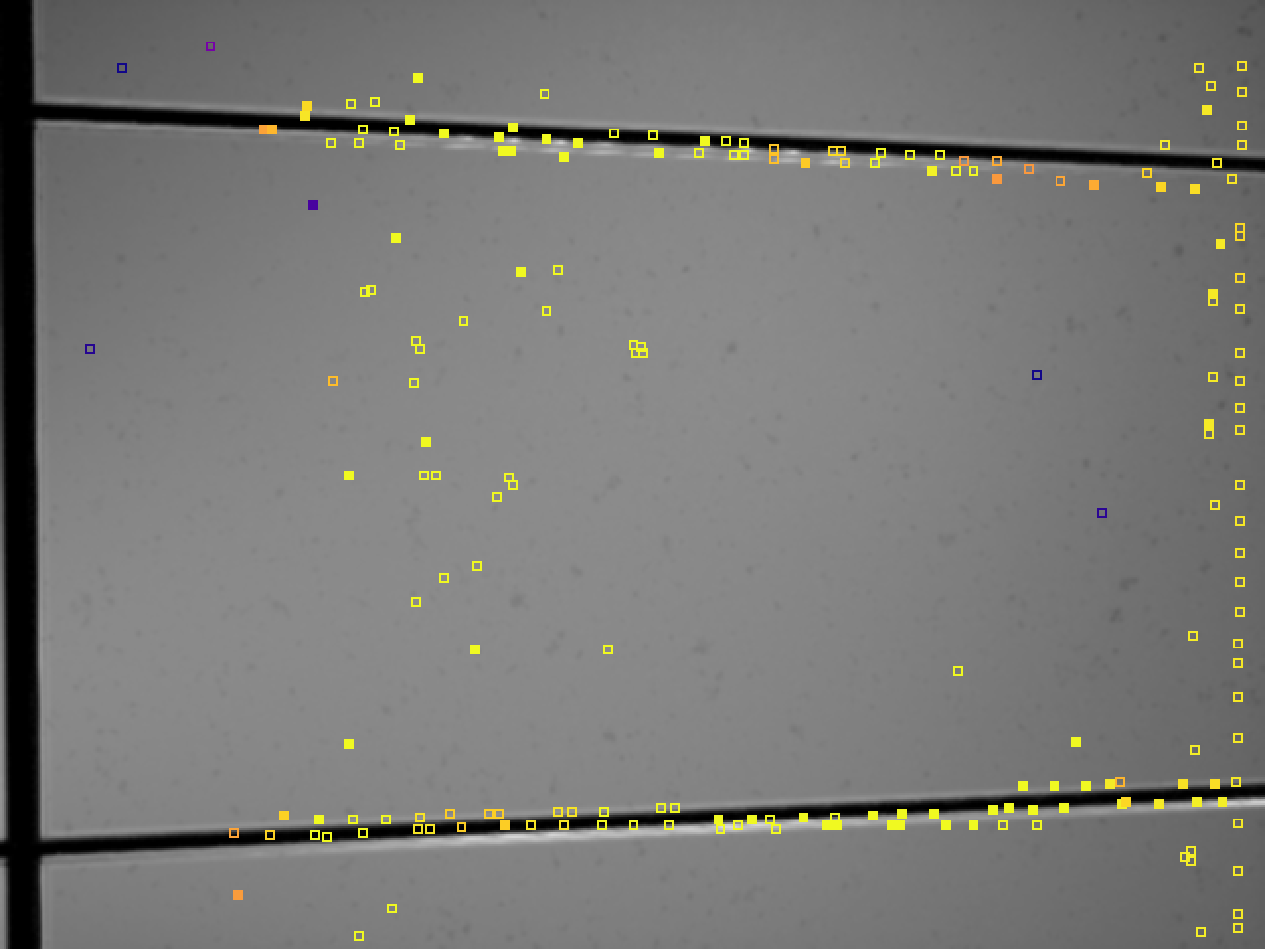}
\end{subfigure}
\caption{
\textbf{Failure cases} of our method on the  TartanAir dataset~\cite{Wang2020TartanAirAD}.
Causes of failure from left to right are: repetitive texture, blocked camera and low texture.
}
\label{fig:result/failure}
\vspace{-5mm}
\end{figure}

We adopt the data-oriented design philosophy used in~\cite{Qu2021LLOLLO}, where we leverage array-based data structures for better cache-locality, ease of parallelism and zero dynamic allocation at runtime.
Each major stage in our pipeline can parallelize its computation, including point selection, stereo matching, frame tracking, and bundle adjustment.
We apply a task-based, fine-grained parallelism, where each task contains a single row of cells defined by the point selection process.
We found that this granularity struck a good balance between context-switching overhead and resource utilization.

For direct image alignment, each task accumulates its own Hessian and the final result is reduced by a fork-join model~\cite{Voss2019ProTBB}. 
For \gls{pba}, due to the $O(N^2)$ complexity induced by pairwise projections, we add an extra keyframe-level parallelization.
Specifically, we spawn tasks to accumulate block Hessians as long as they do not depend on the same set of points. 
The final Hessian is updated by $N(N-1)$ block Hessians, which requires locking to prevent data races.
However, since $N$ is typically small, this mutex has a low contention rate and the performance impact is negligible.

Our implementation follows modern C++ design principles~\cite{stroustrup2014tour}, with no manual resource or thread management.
Under default settings, our system allocate less than 10MB of memory, which fits into the L3 cache of most modern processors. 
The small memory footprint is crucial to deployment on resource constrained platforms.

We use a window of $N=4$ keyframes and a scale pyramid of $P=5$ levels. 
We found that further increasing $N$ has diminishing returns in accuracy since we already form a set of strongly-connected keyframes.
We benchmark against an open-source implementation of \gls{sdso}\footnote{\url{https://github.com/RonaldSun/stereo_DSO}}, which is based on the official \gls{dso} with stereo integration.
Code\footnote{\url{https://github.com/ShreyasSkandanS/stereo_DSO}} and data\footnote{\url{https://tinyurl.com/2z3nzyre}} to reproduce our results are also publicly available.
\section{Evaluation Results}\label{sec:result}

\subsection{Datasets}

\subsubsection{KITTI Odometry}
The KITTI Odometry dataset~\cite{Geiger2013VisionMR} is an urban driving dataset with stereo sequences and associated ground truth trajectories. 
We evaluate on all 11 (00-10) training sequences.

\subsubsection{Virtual KITTI-2}
The VKITTI2 dataset~\cite{Cabon2020VirtualK2} is a synthetic video dataset intended to appear in the same environment as KITTI 
with simulated lighting and weather conditions.
We use all 5 sequences, each with 10 different variations (including challenging ones like \textit{fog} and \textit{rain}). 

\subsubsection{Tartan Air} TartanAir~\cite{Wang2020TartanAirAD} is a synthetic dataset for robot navigation tasks collected in photo-realistic simulation environments with various styles and scenes.
It contains diverse 6\gls{dof} motion patterns, making it more challenging compared to driving datasets.
We hand-picked 6 scenes (carwelding, gascola, oldtown, office, office2, hospital) from the dataset, comprised of a total of 73 sequences covering both indoor, outdoor, static, and dynamic environments. 
We only use the \textit{Easy} variation of each sequence.


\begin{figure}
\vspace{6px}
\centering
\includegraphics[width=\linewidth]{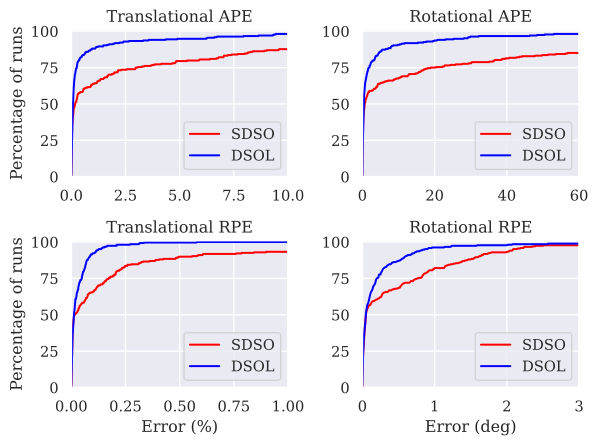}
\caption{
\textbf{Cumulative error plots} of \gls{sdso} vs. \gls{dsol} on all runs (268 total).
y-axis shows percentage of runs.
For an error value $x$, the corresponding $y$ value is the percentage of runs that has smaller error.
} 
\label{fig:result/acc/cumulative}
\vspace{-3mm}
\end{figure}

\begin{figure}
\centering
\includegraphics[width=\linewidth]{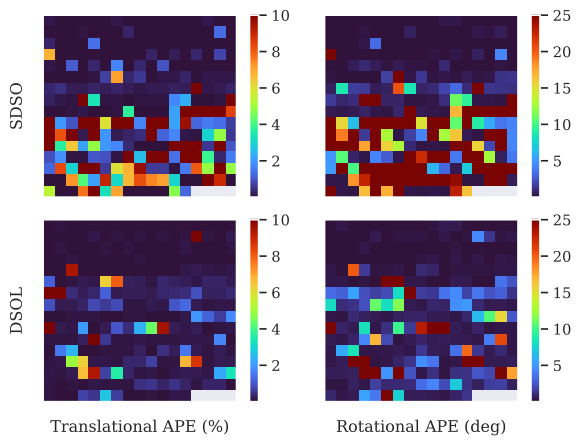}
\caption{
\textbf{Error heatmaps} of Trans- and Rot-APE for all 268 runs.
Each pixel represents one metric of a single run by one method. Trans- and Rot-APE are capped at $10\%$ and $25^\circ$ respectively, which typically indicate gross tracking failure. 
The order of runs from top left to bottom right is VKITTI, KITTI, TartanAir. 
This figure is best viewed in color} 
\label{fig:result/acc/heatmap}
\vspace{-5mm}
\end{figure}

\subsection{Accuracy and Robustness}
\begin{figure*}[t]
\vspace{6px}
\begin{subfigure}[b]{0.262\linewidth}
    \centering
    \includegraphics[width=\linewidth]{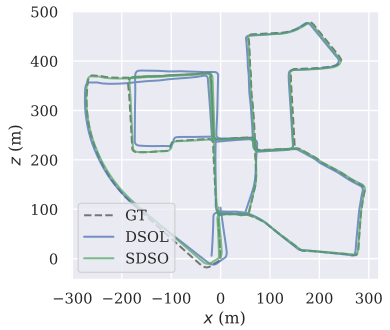}
    \caption{KITTI 00}
\end{subfigure}
\begin{subfigure}[b]{0.255\linewidth}
    \centering
    \includegraphics[width=\linewidth]{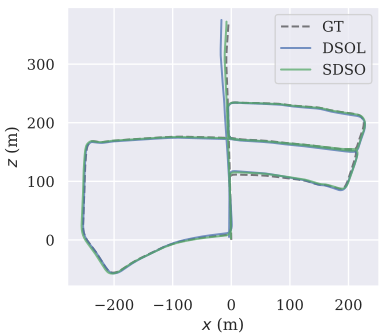}
    \caption{KITTI 05}
\end{subfigure}
\begin{subfigure}[b]{0.275\linewidth}
    \centering
    \includegraphics[width=\linewidth]{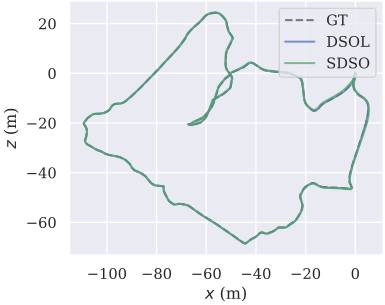}
    \caption{gascola 004}
\end{subfigure}
\begin{subfigure}[b]{0.185\linewidth}
    \centering
    \includegraphics[width=\linewidth]{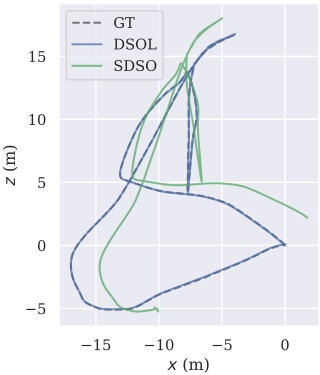}
    \caption{office 004}
\end{subfigure}
\caption{
Qualitative results of \gls{sdso} and \gls{dsol} on selected sequences from KITTI~\cite{Geiger2013VisionMR} and TartanAir~\cite{Wang2020TartanAirAD}.
}
\label{fig:result/evo}
\vspace{-5mm}
\end{figure*}

To evaluate odometry accuracy, we run \gls{sdso} and \gls{dsol} on VKITTI2, KITTI and TartanAir, both forward and backward, resulting in a total of $(50 + 11 + 73) \times 2 = 268$ runs with a wide variety of scenarios.
We report both translational and rotational \gls{ape} and \gls{rpe}, 
computed by the \textit{evo}\footnote{\url{https://github.com/MichaelGrupp/evo}} package. 
\gls{ape} measures the absolute pose error between the reference and the estimated trajectory, whereas \gls{rpe} measures the relative pose error between two consecutive frames. 
We use Umeyama alignment~\cite{umeyama1991least} as a pre-processing step to align the estimation to groundtruth. 
Note that due to the vast scale differences between different sequences, we normalize the translational metrics by the corresponding sequence length to obtain percentage errors.
Both methods are allowed to re-initialize when tracking is lost. 
However, if a method cannot track the full sequence (due to the program crashing),
we consider it a failure only if it tracks no more than $80\%$ of the run and assign it a fixed large RMSE.

Fig.~\ref{fig:result/acc/cumulative} shows the cumulative error plots of \gls{sdso} and \gls{dsol} on all runs.
A point $(x, y)$ on a curve indicates that $y\%$ of runs has error less than $x$.
Therefore a curve closer to the top-left corner has better overall accuracy. 
We see that \gls{dsol} outperforms \gls{sdso} by a large margin, as evidenced by the area between the two curves.

We also show error heatmaps of all runs in Fig.~\ref{fig:result/acc/heatmap}, where each pixel represents the corresponding metric from one run.
The runs are in the order of VKITTI, KITTI, and TartanAir from top left to bottom right within each plot.
We see that both methods perform relatively well on KITTI and VKITTI, where camera motions are smooth and planar. 
However, \gls{sdso} struggles when facing the diverse 6\gls{dof} motion patterns from TartanAir, resulting in frequent tracking failures. 
Our method was able to track most of the sequences from TartanAir with good accuracy and without re-initialization, indicating a higher level of robustness.
However, there are still several failure cases that are generally impossible for a vision-based odometry system, which we visualize in Fig.~\ref{fig:result/failure}.
Qualitative results on selected sequences are shown in Fig.~\ref{fig:result/evo}.

\subsection{Speed}
\begin{table}
\vspace{3px}
\centering
\caption{Runtime (in ms) of \gls{sdso} and \gls{dsol} on different processors
under single and multi-threaded settings.  
Results are averaged among several testing sequences from TartanAir (with stereo images at $640\times 480$ resolution).
S denotes single-thread mode, M-$n$ denotes multi-thread mode with at most $n$ threads.
}
\label{tab:result/runtime}
\begin{tabular}{ccccccc}
\hline
Runtime [ms]  & \multicolumn{2}{c}{E5-2683} & 
\multicolumn{2}{c}{i7-11800H}      & \multicolumn{2}{c}{ARMv8.2} \\
name-stage & S & M-32 & S & M-16 & S & M-6\\
\hline \hline
\gls{sdso}-track  & 13.95 & 15.02 & 7.18  & 7.29  & 34.39 & 32.73 \\
\gls{dsol}-track  & 4.81  & 2.27  & 2.39  & 1.23  & 9.35  & 5.08 \\
Speed up          & 2.90x & 6.67x & 3.00x & 5.93x & 3.68x & 6.44x \\
\hline 
\gls{sdso}-kf     & 79.07 & 52.64 & 71.59 & 40.56 & 282.6 & 195.5\\
\gls{dsol}-kf     & 46.77 & 7.82  & 25.45 & 9.96  & 113.4 & 41.66 \\
Speed up          & 1.69x & 6.73x & 2.81x & 4.07x & 2.49x & 4.69x \\
\hline
\end{tabular}
\vspace{-5mm}
\end{table}

We benchmark \gls{sdso} and \gls{dsol} on various compute units with single- and multi-threaded settings. 
Processor types includes an Intel Xeon E5-2683 (16$\times$3.0GHz), an Intel i7-11800H (8$\times$4.6GHz) and an Nvidia Xavier NX ARMv8.2 (6$\times$1.4GHz), which covers a wide range of core counts and clock rates.
We run both systems with stereo inputs on the gascola scene from the TartanAir dataset and average timing results over several runs. 
We use the default settings for \gls{sdso} (with 7 keyframes and 2000 points max) and do not enforce real-time execution (no skipping frames).
Note that it is difficult to ensure a completely fair comparison, as each system uses slightly different window sizes, pyramid levels, number of iterations and other hyper-parameters that may affect its performance. 

Runtime results are shown in Table~\ref{tab:result/runtime}.
We divide each system into two stages:
\texttt{track} is performed at the arrival of each image, while \texttt{kf} time is only recorded when adding a new keyframe.
\gls{dsol} runs faster than \gls{sdso} in all benchmarks we performed.
It is on average around \textbf{4.8x} faster in frame tracking and \textbf{3.8x} faster in keyframe creation. 
Our system can track frames at more than \textbf{500} \gls{fps} and even handle keyframes at \textbf{100} \gls{fps} on laptop-grade hardware.
It is also worth noting that \gls{sdso} has relatively poor parallel scalability (by dividing numbers under M by S in Table~\ref{tab:result/runtime}), whereas ours can often achieve further speedup where more computation resources are available.

\subsection{High FPS Tracking}

\begin{figure}
\vspace{6px}
\begin{subfigure}[b]{0.495\linewidth}
    \centering
    \includegraphics[width=\linewidth]{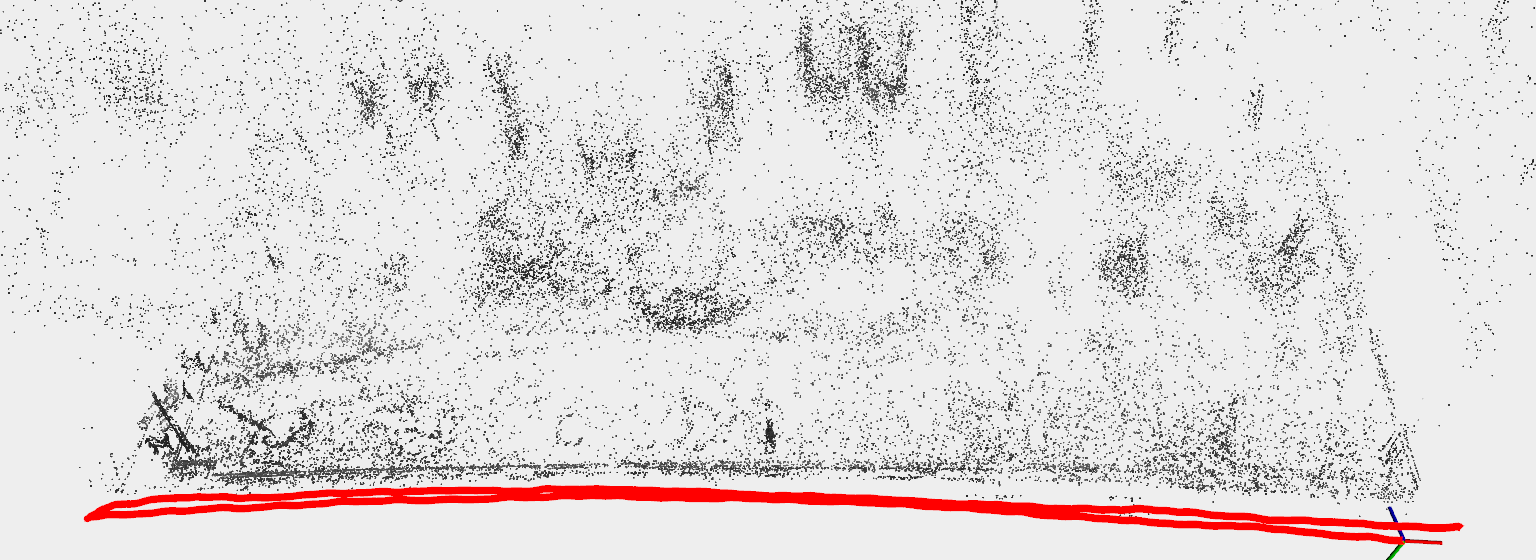}
\end{subfigure}
\hfill
\begin{subfigure}[b]{0.495\linewidth}
    \centering
    \includegraphics[width=\linewidth]{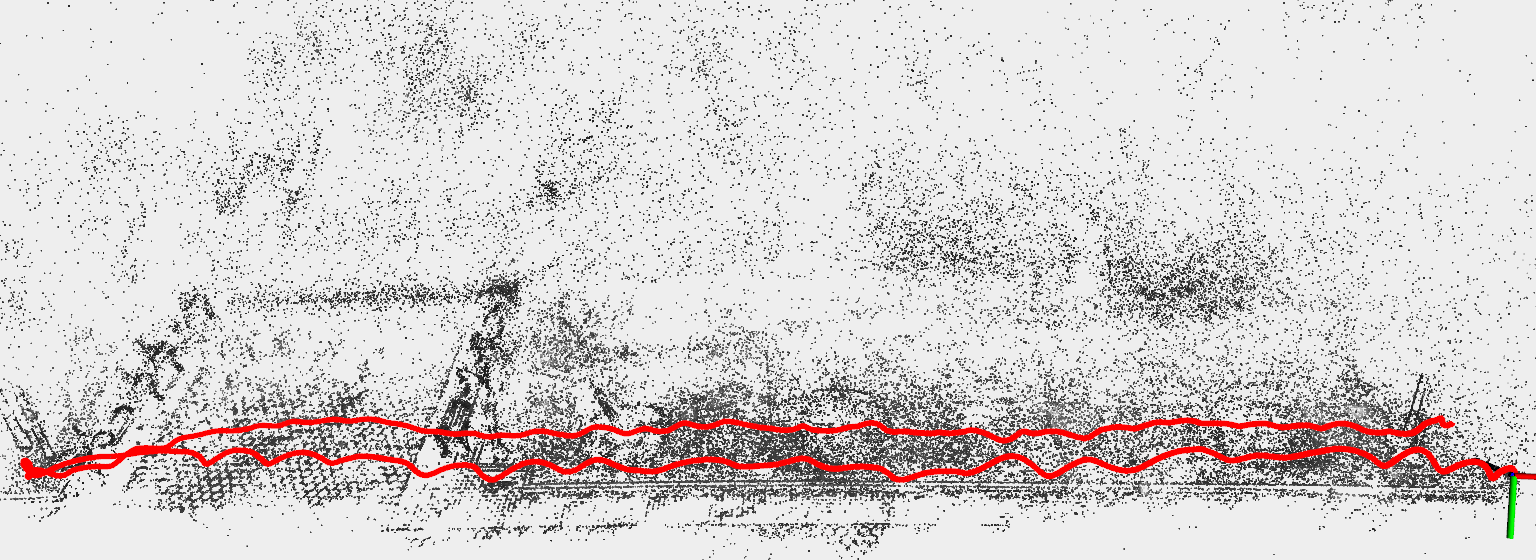}
\end{subfigure}
\caption{Qualitative results of \gls{dsol} on our Realsense D455 dataset with 60Hz stereo images at 640$\times$480 resolution.
}
\label{fig:result/realsense}
\vspace{-5mm}
\end{figure}

As a final experiment, we collected datasets to showcase tracking at high frame rates with the Intel Realsense D455 sensor.
We disabled the IR emitter on the D455 and collected stereo images from the two grayscale cameras at 60\gls{fps} with an image resolution of 640$\times$480.
Qualitative results obtained on two hand-held sequences are show in Fig.~\ref{fig:result/realsense}. 
Our system is able to process the 60Hz stream in real time without dropping any frames.

\section{Conclusion}

In this paper, we have presented a fast, accurate and robust direct sparse odometry (DSOL) that is suitable for robotics applications on resource constrained platforms.
Our system builds on ideas from \gls{dso} and \gls{sdso} with several algorithmic and implementation improvements that enable it to handle high frame rates.
The significant speed boost was also made possible by designing with parallelization in mind, allowing efficient usage of multi-core processors.
Our experiments have shown that our stereo version out-performed the baseline \gls{sdso} on challenging datasets in terms of accuracy and speed.
We open source our implementation as a contribution to the community.



\bibliographystyle{IEEEtran}
\bibliography{IEEEabrv,reference}

\begin{thebibliography}{10}
\providecommand{\url}[1]{#1}
\csname url@samestyle\endcsname
\providecommand{\newblock}{\relax}
\providecommand{\bibinfo}[2]{#2}
\providecommand{\BIBentrySTDinterwordspacing}{\spaceskip=0pt\relax}
\providecommand{\BIBentryALTinterwordstretchfactor}{4}
\providecommand{\BIBentryALTinterwordspacing}{\spaceskip=\fontdimen2\font plus
\BIBentryALTinterwordstretchfactor\fontdimen3\font minus
  \fontdimen4\font\relax}
\providecommand{\BIBforeignlanguage}[2]{{%
\expandafter\ifx\csname l@#1\endcsname\relax
\typeout{** WARNING: IEEEtran.bst: No hyphenation pattern has been}%
\typeout{** loaded for the language `#1'. Using the pattern for}%
\typeout{** the default language instead.}%
\else
\language=\csname l@#1\endcsname
\fi
#2}}
\providecommand{\BIBdecl}{\relax}
\BIBdecl

\bibitem{Engel2018DirectSO}
J.~Engel, V.~Koltun, and D.~Cremers, ``Direct sparse odometry,'' \emph{IEEE
  Transactions on Pattern Analysis and Machine Intelligence}, vol.~40, pp.
  611--625, 2018.

\bibitem{Wang2017StereoDL}
R.~Wang, M.~Schw{\"o}rer, and D.~Cremers, ``Stereo dso: Large-scale direct
  sparse visual odometry with stereo cameras,'' \emph{2017 IEEE International
  Conference on Computer Vision (ICCV)}, pp. 3923--3931, 2017.

\bibitem{MurArtal2015ORBSLAMAV}
R.~Mur-Artal, J.~M.~M. Montiel, and J.~D. Tard{\'o}s, ``Orb-slam: A versatile
  and accurate monocular slam system,'' \emph{IEEE Transactions on Robotics},
  vol.~31, pp. 1147--1163, 2015.

\bibitem{Engel2014LSDSLAMLD}
J.~Engel, T.~Sch{\"o}ps, and D.~Cremers, ``Lsd-slam: Large-scale direct
  monocular slam,'' in \emph{ECCV}, 2014.

\bibitem{Mohta2018FastAF}
K.~Mohta, M.~Watterson, Y.~Mulgaonkar, S.~Liu, C.~Qu, A.~Makineni, K.~Saulnier,
  K.~Sun, A.~Z. Zhu, J.~A. Delmerico, K.~Karydis, N.~A. Atanasov, G.~Loianno,
  D.~Scaramuzza, K.~Daniilidis, C.~J. Taylor, and V.~R. Kumar, ``Fast,
  autonomous flight in gps-denied and cluttered environments,'' \emph{ArXiv},
  vol. abs/1712.02052, 2018.

\bibitem{Handa2012RealTimeCT}
A.~Handa, R.~A. Newcombe, A.~Angeli, and A.~J. Davison, ``Real-time camera
  tracking: When is high frame-rate best?'' in \emph{ECCV}, 2012.

\bibitem{Wang2020TartanAirAD}
W.~Wang, D.~Zhu, X.~Wang, Y.~Hu, Y.~Qiu, C.~Wang, Y.~Hu, A.~Kapoor, and S.~A.
  Scherer, ``Tartanair: A dataset to push the limits of visual slam,''
  \emph{2020 IEEE/RSJ International Conference on Intelligent Robots and
  Systems (IROS)}, pp. 4909--4916, 2020.

\bibitem{Irani1999AllAD}
M.~Irani and P.~Anandan, ``All about direct methods,'' 1999.

\bibitem{Davison2007MonoSLAMRS}
A.~J. Davison, I.~D. Reid, N.~Molton, and O.~Stasse, ``Monoslam: Real-time
  single camera slam,'' \emph{IEEE Transactions on Pattern Analysis and Machine
  Intelligence}, vol.~29, pp. 1052--1067, 2007.

\bibitem{Klein2007ParallelTA}
G.~S.~W. Klein and D.~W. Murray, ``Parallel tracking and mapping for small ar
  workspaces,'' \emph{2007 6th IEEE and ACM International Symposium on Mixed
  and Augmented Reality}, pp. 225--234, 2007.

\bibitem{Strasdat2010ScaleDL}
H.~M. Strasdat, J.~M.~M. Montiel, and A.~J. Davison, ``Scale drift-aware large
  scale monocular slam,'' in \emph{Robotics: Science and Systems}, 2010.

\bibitem{Strasdat2011DoubleWO}
H.~M. Strasdat, A.~J. Davison, J.~M.~M. Montiel, and K.~Konolige, ``Double
  window optimisation for constant time visual slam,'' \emph{2011 International
  Conference on Computer Vision}, pp. 2352--2359, 2011.

\bibitem{Cummins2008FABMAPPL}
M.~J. Cummins and P.~Newman, ``Fab-map: Probabilistic localization and mapping
  in the space of appearance,'' \emph{The International Journal of Robotics
  Research}, vol.~27, pp. 647 -- 665, 2008.

\bibitem{Newcombe2011DTAMDT}
R.~A. Newcombe, S.~Lovegrove, and A.~J. Davison, ``Dtam: Dense tracking and
  mapping in real-time,'' \emph{2011 International Conference on Computer
  Vision}, pp. 2320--2327, 2011.

\bibitem{Engel2013SemidenseVO}
J.~Engel, J.~Sturm, and D.~Cremers, ``Semi-dense visual odometry for a
  monocular camera,'' \emph{2013 IEEE International Conference on Computer
  Vision}, pp. 1449--1456, 2013.

\bibitem{Forster2014SVOFS}
C.~Forster, M.~Pizzoli, and D.~Scaramuzza, ``Svo: Fast semi-direct monocular
  visual odometry,'' \emph{2014 IEEE International Conference on Robotics and
  Automation (ICRA)}, pp. 15--22, 2014.

\bibitem{Jin2003ASA}
H.~Jin, P.~Favaro, and S.~Soatto, ``A semi-direct approach to structure from
  motion,'' \emph{The Visual Computer}, vol.~19, pp. 377--394, 2003.

\bibitem{Matsuki2018OmnidirectionalDD}
H.~Matsuki, L.~von Stumberg, V.~C. Usenko, J.~St{\"u}ckler, and D.~Cremers,
  ``Omnidirectional dso: Direct sparse odometry with fisheye cameras,''
  \emph{IEEE Robotics and Automation Letters}, vol.~3, pp. 3693--3700, 2018.

\bibitem{Schubert2018DirectSO}
D.~Schubert, N.~Demmel, V.~C. Usenko, J.~St{\"u}ckler, and D.~Cremers, ``Direct
  sparse odometry with rolling shutter,'' in \emph{ECCV}, 2018.

\bibitem{Gao2018LDSODS}
X.~Gao, R.~Wang, N.~Demmel, and D.~Cremers, ``Ldso: Direct sparse odometry with
  loop closure,'' \emph{2018 IEEE/RSJ International Conference on Intelligent
  Robots and Systems (IROS)}, pp. 2198--2204, 2018.

\bibitem{Yang2020D3VODD}
N.~Yang, L.~von Stumberg, R.~Wang, and D.~Cremers, ``D3vo: Deep depth, deep
  pose and deep uncertainty for monocular visual odometry,'' \emph{2020
  IEEE/CVF Conference on Computer Vision and Pattern Recognition (CVPR)}, pp.
  1278--1289, 2020.

\bibitem{Engel2016APC}
J.~Engel, V.~C. Usenko, and D.~Cremers, ``A photometrically calibrated
  benchmark for monocular visual odometry,'' \emph{ArXiv}, vol. abs/1607.02555,
  2016.

\bibitem{Bergmann2018OnlinePC}
P.~Bergmann, R.~Wang, and D.~Cremers, ``Online photometric calibration of auto
  exposure video for realtime visual odometry and slam,'' \emph{IEEE Robotics
  and Automation Letters}, vol.~3, pp. 627--634, 2018.

\bibitem{Baker2004LucasKanade2Y}
S.~Baker and I.~Matthews, ``Lucas-kanade 20 years on: A unifying framework,''
  \emph{International Journal of Computer Vision}, vol.~56, pp. 221--255, 2004.

\bibitem{Kerl2013RobustOE}
C.~Kerl, J.~Sturm, and D.~Cremers, ``Robust odometry estimation for rgb-d
  cameras,'' \emph{2013 IEEE International Conference on Robotics and
  Automation}, pp. 3748--3754, 2013.

\bibitem{Zubizarreta2020DirectSM}
J.~A. Zubizarreta, I.~Aguinaga, and J.~M.~M. Montiel, ``Direct sparse
  mapping,'' \emph{IEEE Transactions on Robotics}, vol.~36, pp. 1363--1370,
  2020.

\bibitem{Klose2013EfficientCA}
S.~Klose, P.~Heise, and A.~Knoll, ``Efficient compositional approaches for
  real-time robust direct visual odometry from rgb-d data,'' \emph{2013
  IEEE/RSJ International Conference on Intelligent Robots and Systems}, pp.
  1100--1106, 2013.

\bibitem{Lucas1981AnII}
B.~D. Lucas and T.~Kanade, ``An iterative image registration technique with an
  application to stereo vision,'' in \emph{IJCAI}, 1981.

\bibitem{MurArtal2017ORBSLAM2AO}
R.~Mur-Artal and J.~D. Tard{\'o}s, ``Orb-slam2: An open-source slam system for
  monocular, stereo, and rgb-d cameras,'' \emph{IEEE Transactions on Robotics},
  vol.~33, pp. 1255--1262, 2017.

\bibitem{Hertzberg2008AFF}
C.~Hertzberg and G.~Piazzi, ``A framework for sparse, non-linear least squares
  problems on manifolds,'' 2008.

\bibitem{Li2012ConsistencyOE}
M.~Li and A.~I. Mourikis, ``Consistency of ekf-based visual-inertial
  odometry,'' 2012.

\bibitem{Leutenegger2015KeyframebasedVO}
S.~Leutenegger, S.~Lynen, M.~Bosse, R.~Y. Siegwart, and P.~T. Furgale,
  ``Keyframe-based visual–inertial odometry using nonlinear optimization,''
  \emph{The International Journal of Robotics Research}, vol.~34, pp. 314 --
  334, 2015.

\bibitem{Alismail2016PhotometricBA}
H.~Alismail, B.~Browning, and S.~Lucey, ``Photometric bundle adjustment for
  vision-based slam,'' \emph{ArXiv}, vol. abs/1608.02026, 2016.

\bibitem{Qu2021LLOLLO}
C.~Qu, S.~S. Shivakumar, W.~Liu, and C.~J. Taylor, ``Llol: Low-latency odometry
  for spinning lidars,'' \emph{ArXiv}, vol. abs/2110.01725, 2021.

\bibitem{Voss2019ProTBB}
M.~Voss, R.~Asenjo, and J.~Reinders, \emph{Pro TBB: C++ Parallel Programming
  with Threading Building Blocks}, 1st~ed.\hskip 1em plus 0.5em minus
  0.4em\relax USA: Apress, 2019.

\bibitem{stroustrup2014tour}
\BIBentryALTinterwordspacing
B.~Stroustrup, \emph{A Tour of C++}, ser. Always learning.\hskip 1em plus 0.5em
  minus 0.4em\relax Addison-Wesley, 2014. [Online]. Available:
  \url{https://books.google.com/books?id=-VNsAQAAQBAJ}
\BIBentrySTDinterwordspacing

\bibitem{Geiger2013VisionMR}
A.~Geiger, P.~Lenz, C.~Stiller, and R.~Urtasun, ``Vision meets robotics: The
  kitti dataset,'' \emph{The International Journal of Robotics Research},
  vol.~32, pp. 1231 -- 1237, 2013.

\bibitem{Cabon2020VirtualK2}
Y.~Cabon, N.~Murray, and M.~Humenberger, ``Virtual kitti 2,'' \emph{ArXiv},
  vol. abs/2001.10773, 2020.

\bibitem{umeyama1991least}
S.~Umeyama, ``Least-squares estimation of transformation parameters between two
  point patterns,'' \emph{IEEE Transactions on Pattern Analysis \& Machine
  Intelligence}, vol.~13, no.~04, pp. 376--380, 1991.

\end{thebibliography}

\end{document}